\documentclass[11pt]{article}

\usepackage[letterpaper,margin=1in]{geometry}
\usepackage[T1]{fontenc}
\usepackage{lmodern}
\usepackage{placeins}

\usepackage{amsmath,amssymb,amsthm,mathtools,bm}
\usepackage{booktabs}
\usepackage{siunitx}
\usepackage{float}
\usepackage{microtype}
\usepackage{enumitem}
\usepackage{hyperref}
\usepackage{xcolor}
\usepackage{array}
\usepackage{tikz}
\usepackage{tabularx}
\usetikzlibrary{arrows.meta,positioning,calc,backgrounds,decorations.pathreplacing}
\definecolor{cdelta}{RGB}{31,80,180}
\definecolor{cphi}{RGB}{214,96,20}
\definecolor{cA}{RGB}{20,130,70}
\definecolor{cB}{RGB}{130,50,160}
\definecolor{cgy}{RGB}{125,125,125}
\tikzset{
  vec/.style={-{Stealth[length=2.8mm]}, line width=1.05pt},
  tvec/.style={-{Stealth[length=2.2mm]}, line width=0.72pt},
  dot/.style={circle, fill=black, inner sep=1.25pt},
  odot/.style={circle, draw, fill=white, inner sep=1.1pt, line width=.6pt},
  lab/.style={font=\small},
  slab/.style={font=\footnotesize}
}

\hypersetup{colorlinks=true,linkcolor=black,citecolor=blue,urlcolor=blue,
  pdftitle={When Should Self-Training Be Reversed? Geometry, Entropy, and Sampler Compensation in Generative Models},
  pdfauthor={Keshav Goyal, Suraj Yadav}}

\newtheorem{theorem}{Theorem}
\newtheorem{proposition}[theorem]{Proposition}
\newtheorem{corollary}[theorem]{Corollary}

\theoremstyle{definition}

\newtheorem{remark}[theorem]{Remark}

\DeclareMathOperator{\tr}{tr}

\DeclareMathOperator{\range}{range}

\DeclareMathOperator{\Var}{Var}
\DeclareMathOperator{\KL}{KL}
\DeclareMathOperator{\Log}{Log}

\newcommand{\E}{\mathbb{E}}
\newcommand{\PP}{\mathcal P}
\newcommand{\V}{\mathcal V}
\newcommand{\SPD}{\mathrm{SPD}}
\newcommand{\Normal}{\mathcal N}
\newcommand{\Rdata}{R_{\mathrm{data}}}
\newcommand{\Rsyn}{R_{\mathrm{syn}}}
\newcommand{\Jsam}{\mathcal J_{\kappa}}
\newcommand{\Sigstar}{\Sigma_{\star}}
\newcommand{\whSigma}{\widehat\Sigma}
\newcommand{\dd}{\,\mathrm d}
\newcommand{\rev}{\mathrm{rev}}

\begin{document}
\title{When Should Self-Training Be Reversed?\\
Geometry, Entropy, and Sampler Compensation in Generative Models}

\author{
  Keshav Goyal$^{1,*}$ \qquad Suraj Yadav$^{2,*,\dagger}$\\[8pt]
  {\small $^{1}$Stem Cell Institute, University of Cambridge, United Kingdom}\\
  {\small $^{2}$Hertie Institute for AI in Brain Health, Eberhard-Karls-Universit\"at T\"ubingen, Germany}\\[4pt]
  {\small\texttt{kg603@cam.ac.uk}, \texttt{suraj-prakash.yadav@uni-tuebingen.de}}\\[6pt]
  {\footnotesize $^{*}$Equal contribution. \quad $^{\dagger}$Corresponding author: \texttt{suraj-prakash.yadav@uni-tuebingen.de}}
}
\date{}

\maketitle

\begin{abstract}
Negative extrapolation from self-training, introduced by NEON \cite{alemohammad2025neon}, can
improve generated samples without additional real data: one briefly fine-tunes a model
on samples from its own inference pipeline and then moves its parameters in the opposite
direction. The motivating explanation is that practical samplers are mode seeking, in the
sense that their outputs concentrate on high-density regions of the current model, so
self-training amplifies this bias and reversal corrects it. We show that this sharpening
property alone does not determine the sign. It specifies how the synthetic distribution
differs from the current model, but not where that model lies relative to the data
distribution; the same sharpening transformation can therefore make reversal either
beneficial or harmful to the learned model.

For negative-log-likelihood training with global power tempering as the mode seeking map, the synthetic population
gradient is exactly a R\'enyi-entropy gradient: self-training follows entropy descent and
negative extrapolation entropy ascent, so model correction becomes a directional alignment
question. This entropy direction measures transformation-induced volume change, and on
elliptical families an explicit radial-score statistic decides whether increasing volume
helps, simplifying in the $d$-dimensional Gaussian case to $\tr(\Sigma^{-1}\Sigstar) > d$. %

We then separate this model-side question from the quality of the distribution actually
emitted by an inference procedure. A local compositional theory distinguishes model error,
sampler distortion, the training response, and the sampler response, yielding different
criteria for \emph{model correction} and \emph{sampler compensation}. %
A self-training displacement is therefore best viewed
as a probe of the composed model--sampler--training system, not as a universal estimate of
model error.
\end{abstract}
\section{Introduction}
\label{sec:intro}

Generative models are increasingly trained or adapted using model-generated
data, from synthetic-data augmentation to self-consuming training loops
\cite{alemohammad2024mad,shumailov2024collapse,bertrand2024stability,
gerstgrasser2024collapse,fu2024selfconsuming,gillman2024selfcorrecting,
kazdan2025collapse,zhu2026reflow}. Such feedback can be useful, but it can
also amplify errors already present in the generative pipeline.

Negative Extrapolation from self training (NEON)~\cite{alemohammad2025neon} turns this potential failure mode into a
correction signal. Starting from a base model $\theta_r$, it briefly
fine-tunes on samples from its own inference procedure to obtain $\theta_s$,
then extrapolates in the opposite direction,
\begin{equation}
    \theta_{\rev}(w)
    =\theta_r-w(\theta_s-\theta_r),\qquad w>0.
    \label{eq:reversal-family}
\end{equation}
NEON attributes its empirical gains to mode-seeking inference: the sampler
favors high-density regions, self-training reinforces this bias, and reversal
is argued to correct it.

We show that mode seeking alone does not determine this sign. It describes
how the sampler output differs from the current model, whereas model
correction depends on where that model lies relative to the target. The same
mode-seeking sampler can therefore make reversal beneficial for one base
model and harmful for another. A one-dimensional Gaussian scale family (Section \ref{sec:gaussian-scale-example})
already gives a counterexample to the mode-seeking implication used by
NEON; Appendix~\ref{app:monotone-covariance-gap} identifies the precise gap in its proof. 

We then ask what direction self-training actually records. Suppose we take
the current model distribution $p$ and sharpen it by global power tempering, $q_\alpha(x) = \frac{p(x)^\alpha}{\int p^\alpha}, \alpha>1$ so that regions of higher model density receive relatively more weight.
We then self-train the model on samples from this tempered distribution.
For negative log-likelihood (NLL) training, the resulting synthetic
population gradient is exactly a R\'enyi-entropy gradient: self-training
follows entropy descent and reversal entropy ascent
(Theorem~\ref{thm:renyi-direction}).

This gives the self-training direction a concrete geometric meaning.
If the model distribution is smoothly changed---for example, by increasing
or decreasing its scale---the corresponding entropy change records whether
that change expands or contracts probability mass. Whether
reversing that direction actually moves the model toward the target,
however, depends on the relation between the current model and the target. For Gaussians,this leads to a 
simple covariance criterion $\operatorname{tr}(\Sigma^{-1}\Sigma_\star)>d$ (Corollary \ref{cor:trace}).

The Gaussian case also reveals a finite-data effect: in the proportional
high-dimensional regime (Theorem \ref{thm:mp-unsmoothed}), small empirical covariance eigenvalues are
amplified by $\Sigma^{-1}$, and Marchenko--Pastur theory shows that the
favorable trace criterion becomes systematic rather than two-sided.

There is one further complication. The distribution represented by the model,
$p_\theta$, need not be the distribution that its inference procedure
actually generates because of finite-step sampler, temperature change, guidance
scheme etc. Therefore, emitted distribution $S_\kappa(p_\theta)$
may differ systematically from $p_\theta$. Self-training on generated
samples therefore responds not only to errors in the model, but also to
distortion introduced by the sampler. Reversing that update can consequently
improve generated samples in two fundamentally different ways: it may
\emph{correct the model}, moving $p_\theta$ closer to the target, or it may
\emph{compensate for the sampler}, moving the model in a direction that
counteracts the sampler's distortion even if the model itself becomes worse.
We derive separate first-order criteria for these two mechanisms in Proposition \ref{prop:entropy-sampler-dominance} and isolate
them experimentally using controlled diffusion interventions. In particular,
better generated samples alone are not evidence that reversal has corrected
the underlying model.

\paragraph{Contributions.}
\begin{enumerate}
    \item \textbf{Failure of the mode-seeking sign argument.}
    Mode-seeking inference does not by itself imply the gradient anti-alignment required for beneficial reversal; we give an explicit Gaussian counterexample, show that the same issue persists under diffusion score matching, and identify the gap in NEON's proof.

    \item \textbf{Geometry of the self-training direction.}
    Under NLL training and global power tempering, self-training records a
    R\'enyi-entropy direction. We characterize its volume geometry, derive
    elliptical and Gaussian correction criteria, and identify a
    high-dimensional finite-data mechanism for a systematic favorable sign.

    \item \textbf{Model correction versus sampler compensation.}
    A local compositional theory separates model error, sampler distortion,
    training response, and sampler response, yielding distinct criteria for
    improvement of the represented model and of its emitted samples.
\end{enumerate}

\section{Direction of self-training}
\label{sec:direction}

Let $\mathcal M=\{p_\theta:\theta\in\Theta\}$ be the model family,
$p_\star$ the target distribution, and
$p:=p_{\theta_r}$ a fixed base model.  For an inference configuration
$\kappa$, let $q:=S_\kappa(p)$ denote the distribution actually emitted by the sampler.  The configuration
$\kappa$ may encode solver resolution \cite{song2021ddim,karras2022edm,lu2022dpmsolver},
temperature, guidance \cite{dhariwal2021guidance,hosalimans2022cfg}, truncation, or
score rescaling, so in general $q\neq p$. In diffusion models this distinction is
structural: score approximation and numerical sampling error are separate sources of
discrepancy between the learned score field and the emitted law
\cite{song2021sde,chen2023sampling}.

This distinction gives two different notions of quality that we use
throughout the paper:
\[
    \Rdata(\theta)
      :=\E_{p_\star}[-\log p_\theta(X)],
    \qquad
    \Jsam(\theta)
      :=D\!\left(p_\star,S_\kappa(p_\theta)\right).
\] where
$D$ denotes the distributional discrepancy used to evaluate generated samples
(e.g.\ FID, KID, or sliced-Wasserstein distance). Improving $\Rdata(\theta)$
 need not improve the $\Jsam(\theta)$ or vice-versa.  We refer to these two possibilities as
\emph{model correction} and \emph{sampler compensation}, respectively, and
return to their differences in Section~\ref{sec:framework}.

Synthetic self-training on samples from $q$ instead of $p$ optimizes $\Rsyn(\theta) :=\E_q[-\log p_\theta(X)]$. Let
\[
    r_d:=\nabla\Rdata(\theta_r),
    \qquad
    r_s:=\nabla\Rsyn(\theta_r).
\]
If the local optimizer acts through a positive preconditioner $P$, then $\theta_s-\theta_r=-\eta P r_s+O(\eta^2)$. Hence reversing this displacement decreases the real-data risk to first
order exactly when
\[
    \boxed{\langle r_d,Pr_s\rangle<0.}
\]
Thus whether reversal corrects the model is an alignment question involving
both the sampler-induced direction and the current model error.  We first ask
what direction $r_s$ records.

\subsection{A Gaussian scale example}
\label{sec:gaussian-scale-example}

Consider the one-dimensional Gaussian
scale family  $p_t=\mathcal N(0,e^{2t}), \qquad p_\star=\mathcal N(0,1)$. %
Suppose the sampler globally sharpens the model by power tempering: $q_\alpha(x) \propto p_t(x)^\alpha, \alpha>1$. Since powers of a Gaussian remain Gaussian, $q_\alpha =\mathcal N\!\left(0,\frac{e^{2t}}{\alpha}\right)$. The synthetic distribution is therefore narrower than the current model.
NLL self-training moves the fitted model toward smaller scale, while
reversing that displacement moves it toward larger scale.

The population real-data risk is $R_{\mathrm{data}}(t) = t+\frac12e^{-2t}+C$ and hence $R_{\mathrm{data}}'(t)=1-e^{-2t}$. Consequently,
\[
    \boxed{
    \begin{array}{ccl}
        t<0 &:& \text{reversal helps},\\[2mm]
        t>0 &:& \text{reversal hurts}.
    \end{array}}
\]

The sampler is the same in the two cases: it always sharpens the current
model. What changes is the location of the base model relative to the
target. 

\paragraph{Relation to NEON.}
NEON obtains its risk-reduction guarantee in
two steps: \cite[Theorem 1]{alemohammad2025neon} gives anti-alignment conditional on a negative
sampler--error angle, while  \cite[Theorem 2]{alemohammad2025neon} claims that mode-seeking
reweighting forces this angle to be negative. The Gaussian example
above violates the latter implication: power tempering is mode seeking,
yet reversal hurts whenever $t>0$. Appendix~\ref{app:monotone-covariance-gap}
identifies the precise step in the proof of Theorem~2 that fails and
gives the counterexample in NEON's notation. The same sign failure also persists under diffusion score matching; see
Appendix~\ref{app:score-gaussian}.

\subsection{Power tempering and the entropy direction}
\label{sec:renyi}

The Gaussian example shows that power tempering induces a definite
self-training direction, but does not tell us what that direction is for a
general model family. Under NLL training, it has an exact characterization. Power
transformations of densities are standard in tempering and annealing \cite{neal2001ais};
here they serve as an analytically transparent sampler map.

Fix the base model $p=p_{\theta_r}$ and let the synthetic distribution be
\[
    q_\alpha(x)
    =
    \frac{p(x)^\alpha}{Z_\alpha},
    \qquad
    Z_\alpha=\int p(x)^\alpha \dd x,
    \qquad
    \alpha>1.
\]
Recall that the R\'enyi entropy of order $\alpha$ \cite{renyi1961} is
\[
    H_\alpha(p)
    =
    \frac{1}{1-\alpha}
    \log\int p(x)^\alpha\dd x .
\]

\begin{theorem}[Power tempering records a R\'enyi-entropy direction]
\label{thm:renyi-direction}
Under the usual regularity conditions allowing differentiation under the
integral sign,
\[
    \nabla\Rsyn(\theta_r)
    =
    \frac{\alpha-1}{\alpha}\,
    \nabla_\theta H_\alpha(p_\theta)\Big|_{\theta=\theta_r}.
\]
Hence self-training on power-tempered samples follows R\'enyi-entropy
descent, while negative extrapolation follows R\'enyi-entropy ascent.
\end{theorem}

\begin{proof}
Let $u_\theta=\nabla_\theta\log p_\theta$. Since the synthetic
distribution is held fixed during self-training,
\[
    \nabla\Rsyn(\theta_r)
    =
    -\E_{q_\alpha}[u_{\theta_r}]
    =
    -\frac{1}{Z_\alpha}
      \int p_{\theta_r}^\alpha u_{\theta_r}\dd x .
\]
On the other hand,
\[
    \nabla_\theta H_\alpha(p_\theta)
    =
    \frac{\alpha}{1-\alpha}\,
    \frac{\int p_\theta^\alpha u_\theta\dd x}
         {\int p_\theta^\alpha\dd x},
\]
which gives the result.
\end{proof}

This answers the first question raised in Section~\ref{sec:gaussian-scale-example}:
power tempering determines an \emph{entropy direction}. It does not,
by itself, determine whether that direction points toward the target.
The Gaussian example already showed that the same entropy-increasing
reversal can help or hurt depending on the current model.

The result is not specific to power tempering. Appendix~\ref{app:density-reweighting}
gives the more general density-reweighting theory
$q\propto\omega(p)$, for which NLL self-training is generated by an
associated entropy-like potential.
\begin{remark}
Theorem~\ref{thm:renyi-direction} identifies the synthetic gradient, but
not the parameter displacement produced by an arbitrary optimizer.  If the local update acts through a positive preconditioner $P$, the reversal direction is proportional to $P\nabla_\theta H_\alpha$ and can therefore depend on both $P$ and the parametrization.
\end{remark}

\section{Geometry of Entropy direction}
\label{sec:elliptical}

Section~\ref{sec:renyi} identifies the entropy gradient recorded by
power-tempered self-training. Under any positive preconditioner, reversal
is entropy-increasing to first order, but the preconditioner determines
which entropy-increasing direction is taken. We now ask what this entropy
geometry means along explicit deformations of the model distribution, and
when such a deformation reduces population NLL. These are geometric
directional criteria; they do not assert that an arbitrary optimizer
produces the same distribution-space tangent.

Let $p_t=(T_t)_\#p_0$ be a smooth deformation with velocity field $v_t$. Then
\begin{equation}
\label{eq:transport-entropy}
    \frac{d}{dt}H_\alpha(p_t)
    =
    \E_{q_{\alpha,t}}\!\left[\nabla\!\cdot v_t\right],
    \qquad
    q_{\alpha,t}\propto p_t^\alpha .
\end{equation}
Thus power tempering records the local volume change of the deformation, weighted
toward the high-density regions favored by the sampler. In particular,
volume-preserving directions are invisible. The derivation and regularity
conditions are given in Appendix~\ref{app:transformation-families}.

For the affine family
\[ p_{\mu,A}(x)=|\det A|^{-1}g\!\left(A^{-1}(x-\mu)\right),
\]
\eqref{eq:transport-entropy} reduces to
\[
    H_\alpha(p_{\mu,A})=H_\alpha(g)+\log|\det A|.
\]
Hence translations and determinant-preserving shape changes carry no entropy
signal. The appearance of the log-determinant is the familiar change-of-variables volume
term from transformation-based generative models and normalizing flows
\cite{papamakarios2021flows}. For isotropic scaling, $A=e^tI$, one simply has
$H_\alpha(p_t)=H_\alpha(g)+dt$: self-training under sharpening contracts scale,
while reversal expands it, independently of the base shape $g$.

The same observation gives an explicit model-correction criterion for the
isotropic scale direction in a fixed-shape elliptical family \cite{fang1990elliptical}
\[
    p_\Sigma(x)
    =
    c_f|\Sigma|^{-1/2}
    f\!\left(x^\top\Sigma^{-1}x\right),
\]
for which
\[
    H_\alpha(p_\Sigma)
    =
    C_{\alpha,f}
    +\frac12\log\det\Sigma .
\]
Define the radial score
\[
    \psi_f(u):=-u\,\frac{d}{du}\log f(u).
\]

\begin{proposition}[Elliptical radial-score criterion]
\label{prop:elliptical-risk}
Along the isotropic expansion $\Sigma(a)=e^a\Sigma$,
\[
    \left.\frac{d}{da}\Rdata(\Sigma(a))\right|_{a=0}
    =
    \frac d2-
    \E_{p_\star}\!\left[\psi_f(X^\top\Sigma^{-1}X)\right].
\]
Therefore isotropic entropy-increasing reversal improves population NLL exactly when
\[
    \E_{p_\star}\!\left[\psi_f(X^\top\Sigma^{-1}X)\right]>\frac d2 .
\]
\end{proposition}

A proof is given in Appendix~\ref{app:elliptical-proof}.

The sampler therefore determines a log-volume direction, while the target
determines whether that direction is corrective. For a Gaussian model,
$\psi_f(u)=u/2$, giving:

\begin{corollary}[Gaussian trace criterion]
\label{cor:trace}
If the model is $\mathcal N(0,\Sigma)$ and the target is any zero-mean
distribution with finite covariance $\Sigstar$, then
\[
    \boxed{
    \text{isotropic reversal improves NLL}
    \quad\Longleftrightarrow\quad
    \tr(\Sigma^{-1}\Sigstar)>d.}
\]
\end{corollary}
More generally, let $V$ be any symmetric covariance tangent and define
\[
    W:=\Sigma^{-1/2}V\Sigma^{-1/2},
    \qquad
    M:=\Sigma^{-1/2}\Sigstar\Sigma^{-1/2}.
\]
Direct differentiation gives
\[
    D\Rdata(\Sigma)[V]
    =
    \frac12\tr\!\bigl[W(I-M)\bigr].
\]
Thus a positive displacement along $V$ improves population NLL exactly when
\[
    \tr\!\bigl[W(M-I)\bigr]>0.
\]
Corollary~\ref{cor:trace} is the isotropic special case $V=\Sigma$, for
which $W=I$.
A complementary Fisher--Rao criterion: expansion points toward the target iff
$\det\Sigma<\det\Sigstar$ is proved in
Appendix~\ref{app:gaussian-fisher-sign} (Proposition~\ref{prop:det-criterion}).

\subsection{Model correction in finite-data Gaussians}
\label{sec:finite-data-theory}

The Gaussian criteria of the previous section become especially informative once the model itself is learned from finite data. We now ask whether statistical estimation creates a systematic sign for reversal, or whether the sign remains essentially random near the optimum. The answer depends sharply on the relation between dimension and sample size. This is
closely related to classical high-dimensional covariance estimation, where the sample
covariance becomes poorly conditioned and shrinkage toward the identity is a standard
regularization strategy \cite{ledoitwolf2004}. Full proofs and finite-sample calculations
are given in Appendix~\ref{app:wishart}.

\subsection{Classical and proportional regimes}
\label{sec:wishart-main}

Without loss of generality assume $\Sigstar=I$ and let $S_n=n^{-1}\sum_{i=1}^nX_iX_i^\top$. The behavior depends on how dimension scales with sample size.

\begin{proposition}[Fixed-dimensional regime]
\label{prop:fixed-d-wishart}
For Gaussian samples and fixed $d$,
\begin{equation}
    \sqrt n\,\log\det S_n\Rightarrow N(0,2d),
    \qquad
    \sqrt n\,[\tr(S_n^{-1})-d]\Rightarrow N(0,2d).
    \label{eq:fixed-d-clt}
\end{equation}
Consequently both local isotropic-scale criteria remain asymptotically
two-sided: their favorable probabilities tend to $1/2$ as $n\to\infty$.
\end{proposition}

Thus, in fixed dimension, concentration toward the optimum does not by itself create a deterministic reversal sign, since the leading fluctuation remains centered. A two-dimensional Monte Carlo illustration of this finite-sample behavior is given in Appendix~\ref{app:wishart}.

The conclusion changes when $d$ grows with $n$. We use the standard
Marchenko--Pastur asymptotics for sample covariance matrices
\cite{marchenko1967,baisilverstein2010}.

\begin{theorem}[Proportional high-dimensional regime]
\label{thm:mp-unsmoothed}
Suppose $d,n\to\infty$ with $d/n\to\gamma\in(0,1)$ and the whitened sample covariance satisfies the Marchenko--Pastur law together with the corresponding inverse-moment and log-determinant convergence. Then almost surely
\[
    \frac1d\tr(S_n^{-1})\longrightarrow\frac1{1-\gamma}>1,
\]
and
\[
    \frac1d\log\det S_n
    \longrightarrow
    \frac{1-\gamma}{\gamma}\log\frac1{1-\gamma}-1<0.
\]
Thus the isotropic Gaussian-model NLL criterion is favorable with probability tending to one; when the target is Gaussian, the Fisher determinant criterion is favorable as well.
\end{theorem}

The inverse-eigenvalue margin is $\gamma/(1-\gamma)$. Hence high-dimensional estimation systematically creates collapsed directions that receive large weight in $\tr(S_n^{-1})$. This is a statistical mechanism by which the canonical isotropic
entropy-increasing direction becomes genuinely corrective, distinct from
sampler compensation as defined in Section~\ref{sec:framework}.

The Gaussian model therefore serves as a useful toy case of the broader theory: transformation geometry determines the synthetic volume direction, the radial score determines model-risk alignment, and the quadratic radial score makes random-matrix theory available.
\begin{remark}[Smoothing phase boundary]
\label{rem:mp-smoothing}
Continue in the whitened setting $\Sigstar=I$ of
Theorem~\ref{thm:mp-unsmoothed}. To see how additional isotropic smoothing can change the sign, let
$\Sigma_h=S_n+h^2I$. Then under the assumptions of Theorem~\ref{thm:mp-unsmoothed}, isotropic
expansion of $\Sigma_h$ improves Gaussian population NLL asymptotically
if and only if
\[
    h^2<\frac{\gamma}{1+\gamma}.
\]
At equality the first-order scale derivative vanishes, while above this
threshold smoothing dominates the finite-sample spectral collapse and
the same expansion worsens model risk. The proof is given in
Appendix~\ref{app:wishart}.
\end{remark}
\section{Model correction versus sampler compensation}
\label{sec:framework}

\label{sec:framework-local}
Recall the notation from Section \ref{sec:direction}:
At the base model define $\delta:=p-p_\star$ and $\varphi:=q-p=S_\kappa(p)-p$.
Thus the sampler output has residual $q-p_\star=\delta+\varphi$.
The decomposition is exact: $\delta$ is the error already present in the
model, while $\varphi$ is the additional displacement introduced by the
chosen inference procedure.
Now self-train the base model briefly on samples from $q$, obtaining
$\theta_s$.  Let $a$ denote the first-order change in the \emph{model
distribution} produced by this self-training step.  Negative extrapolation by
weight $w>0$ then moves the model in the opposite direction,
\begin{equation}
    p_{\rm rev}(w)=p-wa+o(w).
    \label{eq:model-reversal-a}
\end{equation}
Passing this perturbed model through the same sampler produces, to first
order,
\begin{equation}
    S_\kappa(p_{\rm rev}(w))
      =q-wb+o(w),
    \label{eq:sample-reversal-b}
\end{equation}
where $b$ is the corresponding change in the \emph{emitted distribution}.
There is no reason in general for $b=a$: the sampler acts on the perturbed
model before samples are observed. This is summarized in Table \ref{tab:composition}.
The self-training
experiment records a model-side direction $a$, but sample quality responds to
the sampler-side direction $b$.  Consequently, improvement of
$S_\kappa(p_{\rm rev})$ does not by itself imply improvement of
$p_{\rm rev}$.

A formal tangent-space treatment of these perturbations, including their
relation to relative-density coordinates and the Fisher metric, is given in
Appendix~\ref{app:operators}.  In that notation one may write
$a=A\varphi$ and $b=B_\kappa A\varphi$, but these are not
needed for the main argument below.

For population NLL, a small reversal improves the model distribution exactly
when its first directional derivative is negative.  Using the Fisher metric
\[
    g_p^F(h,k)
    :=
    \int \frac{h(x)k(x)}{p(x)}\,\dd x,
\]
defined formally in
Appendix~\ref{app:operators}, this criterion becomes
\begin{equation}
    \boxed{
    \text{model improves}
    \quad\Longleftrightarrow\quad
    g_p^F(\delta,a)>0.
    }
    \label{eq:model-improvement-criterion}
\end{equation}

For emitted samples, we regard the sample discrepancy as a
functional of the emitted distribution $\mathcal J(r):=D(p_\star,r)$. %
If $\mathcal J$ is differentiable at $q$, then a small reversal improves
sample quality exactly if and only if $\mathrm d\mathcal J_q[b]>0$ where $\mathrm d\mathcal J_q[b]$ denotes the directional derivative of
$\mathcal J$ at $q$ in direction $b$; the general formulation is given in
Appendix~\ref{app:two-risk-proofs}.

If $\mathcal J$ is locally quadratic near $p_\star$, with local bilinear
form $G_J$, this reduces, up to higher-order terms, to
\begin{equation}
    \boxed{
    \text{samples improve}
    \quad\Longleftrightarrow\quad
    G_J(\delta+\varphi,b)>0.
    }
    \label{eq:sample-improvement-criterion}
\end{equation}

Equations~\eqref{eq:model-improvement-criterion} and
\eqref{eq:sample-improvement-criterion} are the key distinction.  The model
criterion is measured against $\delta$, whereas the sample criterion is
measured against the emitted residual $\delta+\varphi$ and the sampler-visible
direction $b$.  They can therefore have opposite signs.

We use \emph{model correction} when reversal improves the underlying model,
and \emph{sampler compensation} when reversal improves emitted samples even
though the underlying model worsens.  %
A useful sufficient condition for the sampler contribution to dominate the
sample criterion is
\begin{equation}
    G_J(\varphi,b)
      >
    \left|G_J(\delta,b)\right|.
    \label{eq:sampler-dominance-sufficient}
\end{equation}
When this holds together with $g_p^F(\delta,a)<0$, the improvement in generated samples
is genuinely driven by compensation for sampler error rather than correction
of model error.

The general directional-derivative statements and their relation to the
operator formulation are proved in Appendix~\ref{app:two-risk-proofs}.

We now illustrate the above discussion via a one-dimensional example. We identify the local coordinate with
$\mathbb R$, and suppose the self-training and sampler responses preserve
orientation:
\[
    a=\rho\varphi,
    \qquad
    b=\beta a,
    \qquad
    \rho,\beta>0.
\]
Then the positive factors $\rho$ and $\beta$ do not affect the signs, and the
two criteria reduce to

\begin{equation}
    \boxed{
    \begin{aligned}
        \text{model improves}
        &\iff \delta\varphi>0,\\
        \text{samples improve}
        &\iff \varphi(\delta+\varphi)>0.
    \end{aligned}}
    \label{eq:scalar-criteria}
\end{equation}

\begin{figure*}[t]
\centering
\resizebox{\textwidth}{!}{%
\begin{tikzpicture}[x=1cm,y=1cm]

\newcommand{\regime}[7]{%
\begin{scope}[xshift=#1cm]
  \draw[black!30] (-2.2,1.0) -- (2.35,1.0);
  \draw[black!30] (-2.2,0.0) -- (2.35,0.0);
  \node[slab,cgy,anchor=south east] at (-2.28,0.95) {model};
  \node[slab,cgy,anchor=south east] at (-2.28,-0.05) {emitted};

  \draw[black!60,dotted,line width=.9pt] (0,-0.55) -- (0,1.75);
  \node[slab,anchor=south] at (0,1.78) {$p_\star$};

  \draw[{Bar[width=4pt]}-{Stealth[length=2mm]},
        cdelta,line width=.8pt]
      (0,1.42) -- (#2,1.42)
      node[midway,above=-1pt,slab,cdelta] {$\delta$};

  \draw[{Bar[width=4pt]}-{Stealth[length=2mm]},
        black!65,line width=.8pt]
      (0,-0.42) -- (#3,-0.42)
      node[midway,below=-1pt,slab,black!65]
      {$\delta+\varphi$};

  \draw[vec,cphi] (#2,1.0) -- (#3,0.0);
  \node[slab,cphi]
      at ({(#2+#3)/2 + 0.34},0.55)
      {$\varphi$};

  \draw[tvec,cA,dashed]
      (#2,1.0) -- ({#2-#4},1.0);
  \draw[tvec,cB,dashed]
      (#3,0.0) -- ({#3-#5},0.0);

  \node[odot,draw=cA] at ({#2-#4},1.0) {};
  \node[odot,draw=cB] at ({#3-#5},0.0) {};
  \node[dot,fill=cdelta] at (#2,1.0) {};
  \node[dot] at (#3,0.0) {};

  \node[slab,anchor=north,align=center]
      at (0.1,-0.92) {#6};
  \node[slab,anchor=north,align=center]
      at (0.1,-1.42) {#7};
\end{scope}}

\regime{0}{1.0}{2.2}{0.70}{0.82}{(a) aligned errors}%
  {$\delta\varphi>0$, $(\delta+\varphi)\varphi>0$\\
   model \textbf{improves}, samples \textbf{improve}}

\regime{6.0}{-0.6}{1.0}{0.72}{0.90}{(b) sampler dominance}%
  {$\delta\varphi<0$, $(\delta+\varphi)\varphi>0$\\
   model \textbf{worsens}, samples \textbf{improve}}

\regime{12.0}{-1.6}{-1.0}{0.38}{0.30}{(c) sampler too weak}%
  {$\delta\varphi<0$, $(\delta+\varphi)\varphi<0$\\
   model \textbf{worsens}, samples \textbf{worsen}}

\end{tikzpicture}}
\caption{The three scalar sign regimes of~\eqref{eq:scalar-criteria}.
The upper track shows the model residual $\delta$ and the lower track the
emitted residual $\delta+\varphi$; the orange segment is the sampler
displacement $\varphi$.  Dashed arrows show the reversed displacement in the
model and after sampling.  Panel (b) is the identifying sampler-compensation
regime: the model moves farther from the target while the emitted samples move
closer.}
\label{fig:regimes}
\end{figure*}
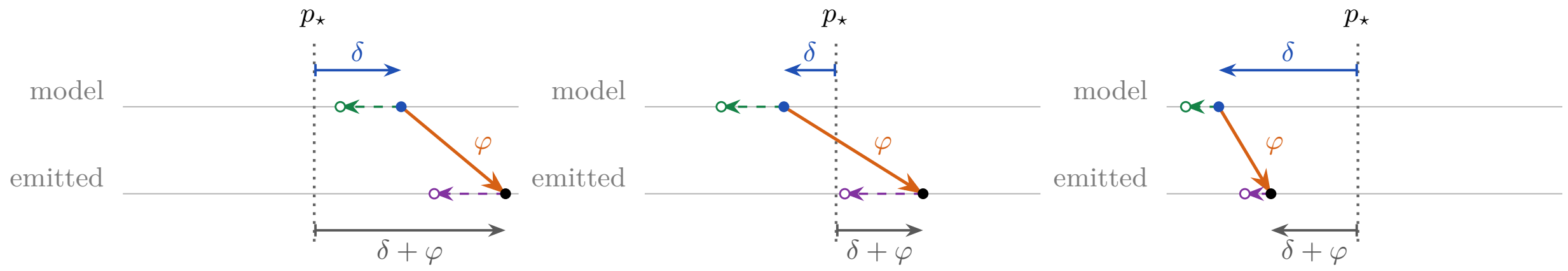

This yields three regimes which are shown and explained in Figure \ref{fig:regimes}

\subsection{From entropy ascent to sampler compensation}
\label{sec:entropy-sampler-dominance}

Section~\ref{sec:renyi} showed that, under power tempering, reversal
follows a R\'enyi-entropy ascent direction.  Figure \ref{fig:regimes}
showed that improving the model and improving the emitted samples are
nevertheless different questions.

\begin{proposition}[Power-tempering ordering]
\label{prop:entropy-sampler-dominance}
Let
\[
    q=S_\alpha(p)=\frac{p^\alpha}{Z_\alpha},
    \qquad
    L=p^{\alpha-1},
    \qquad
    \alpha>1,
\]
and let $\varphi=q-p$. Assume that the sampler-induced displacement is
locally representable by the model $\varphi\in T_p\mathcal M$
and that the local NLL training response preserves this direction, $a=\rho\varphi$ and $\rho>0$.
Use the same power-tempering map to evaluate the reversed model. 
 Then:
\begin{enumerate}
\item \emph{Model correction.}
A sufficiently small positive reversal improves the model NLL if and only if
\[
    \E_{p_\star}L<\E_pL.
\]

\item \emph{Emitted-sample improvement.}
A sufficiently small positive reversal improves the NLL of the power-tempered
emitted law if and only if
\[
    \E_{p_\star}L<\E_qL.
\]

\item \emph{Threshold ordering.}
The two thresholds satisfy
\[
    \E_qL-\E_pL
    =
    \frac{\operatorname{Var}_p(L)}{Z_\alpha}
    \ge0,
\]
and hence
\[
    \E_pL\le\E_qL.
\]
\end{enumerate}
\end{proposition}

A proof is given in Appendix~\ref{app:entropy-sampler-dominance}.
Appendix~\ref{app:density-reweighting} extends the potential construction to
general density reweightings $q\propto\omega(p)$, while
Appendix~\ref{app:entropy-sampler-dominance} gives the corresponding
entropy--Fisher identity and the Gaussian trace--determinant relation.
Further remarks on the sampler Jacobian, reachable composed family, and
residual interpretation are collected in
Appendix~\ref{app:sampler-response}.

\section{Experiments}
\label{sec:experiments}

Section~\ref{sec:framework} separates the model residual $\delta$, sampler
distortion $\varphi$, the response recorded by self-training $A\varphi$, and
the sampler-visible response $B_\kappa A\varphi$.  Our experiments test these
objects separately rather than infer mechanism from sample improvement alone.
All local effects are measured along $\theta(w)=\theta_r-wd$ at $w=0$;
negative derivatives mean improvement.  Full protocols, controls, and
uncertainty estimates are in Appendix~\ref{app:experimental-details}.

\subsection{Model position and the correction sign}
\label{sec:model-position-exp}
We first test the Gaussian trace prediction without changing the self-training
procedure.  In a paired 30-seed diffusion experiment, narrow and wide models
are trained from the same target draws and initializations, using covariance
biases $0.8\Sigma_\star$ and $1.2\Sigma_\star$.  The independently measured
trace statistic
$T=\operatorname{tr}(C^{-1}\Sigma_\star)/2$ is $1.290\pm0.022$ and
$0.863\pm0.014$, respectively, placing every pair on opposite sides of the
criterion $T=1$.  The same score-scaled self-training procedure produces a
covariance-expanding reversal in both arms.  Under neutral evaluation,
reversal improves Gaussian-moment KL and held-out denoising risk in all
30 narrow models, but worsens both in all 30 wide models. Thus changing only the base-model position reverses the sign under the
same score-scaled self-training procedure.  This is consistent with,
rather than a direct test of, the isotropic trace criterion in
Corollary~\ref{cor:trace}, since the learned reversal tangent is not
constrained to be purely isotropic.

\subsection{Isolating sampler distortion }
\label{sec:distortion-response-main}
The main mechanism study uses 30 independently trained Gaussian and 30
independently trained bimodal diffusion models \cite{ho2020ddpm,song2021sde}, each
fitted to 30,000 target observations.  Write $q_{\theta,\kappa}=Q_{K,\zeta}[s_\theta]$ and use the same
network with a refined neutral sampler,
$p_\theta^{\mathrm{ref}}=Q_{2000,1}[s_\theta]$, as an operational reference.
Keeping $\theta$ fixed therefore measures
$D(q_{\theta,\kappa},p_\theta^{\mathrm{ref}})$ rather than total target error.
Neutral refinement from $K=20$ to $500$ reduces median Fourier distortion by
$99.91\%$ for both targets, whereas nonneutral score scaling leaves significant
distortion in every one of the 240 base/source conditions after Holm
correction \cite{holm1979}.

To test what training records, we subtract the neutral-reference response:
\[
 \Delta g_\kappa=g_{\rm syn}(q_{\theta,\kappa})
                 -g_{\rm syn}(p_\theta^{\rm ref}),\qquad
 \Delta d_\kappa=d_\kappa-d_{\rm ref}.
\]
The sampler-induced gradient is nonzero and its descent direction aligns with
the independently recorded finite-Adam contrast in all 240 conditions after
Holm correction.  At $\zeta=1.1$, median
$\cos(-\Delta g_\kappa,\Delta d_\kappa)$ is $0.72$--$0.75$.  Replacing the
learned score by the exact target score gives centered gradient directions
with median cosine $0.996$--$0.998$ to the learned-score contrasts.  Hence a
large component of the recorded response is reproducibly induced by inference
distortion itself.

\subsection{Evidence for sampler compensation}
\label{sec:local-sign-main}
We next keep the ordinary NEON displacement $d_\kappa$ fixed and compare the
same discrepancy before and after the deployed sampler:
\[
\begin{aligned}
 r_D&=\left.\frac{d}{dw}D(p_\star,p^{\rm ref}_{\theta(w)})\right|_0,\\
 j_D&=\left.\frac{d}{dw}D(p_\star,q_{\theta(w),\kappa})\right|_0.
\end{aligned}
\]
Both Fourier-feature discrepancy and sliced-Wasserstein squared
\cite{bonneel2015sliced} are used.
At $\zeta=1.1$, deployed quality improves in all 30 bases for every row of
Table~\ref{tab:local-sign-main}; nevertheless, in 16--17 bases the refined
reference output significantly worsens under both metrics.  The four-way
joint test requires $r_F,r_W>0$ and $j_F,j_W<0$, with Holm correction across
all 240 raw base/source conditions.  This directly realizes sampler
compensation relative to the declared reference, without comparing denoising
risk to a different sample objective.

\begin{table}[t]
\centering
\caption{Ordinary reversal at $\zeta=1.1$.  ``Compensation'' requires
reference deterioration and deployed improvement under both objectives after
Holm correction.  Each row uses 30 independently trained bases.}
\label{tab:local-sign-main}
\small
\begin{tabular}{@{}lrrr@{}}
\toprule
Target & $K$ & Deployed improves & Compensation \\
\midrule
Gaussian & 20  & 30/30 & 16/30 \\
Gaussian & 500 & 30/30 & 16/30 \\
Bimodal  & 20  & 30/30 & 16/30 \\
Bimodal  & 500 & 30/30 & 17/30 \\
\bottomrule
\end{tabular}
\end{table}

\paragraph{Changing only evaluation.}
A separate 30-seed Gaussian intervention holds the base model and NEON
direction fixed, keeps evaluation score scaling neutral, and changes only
sampler resolution.  Refining from $K=10$ to $980$ changes the mean local
Gaussian-moment KL slope from $-0.0301$ to $+0.00705$ and switches every model
from help to harm; sliced Wasserstein gives the same endpoint sign switch.
The held-out denoising-risk derivative, which does not use the evaluation
sampler, is positive in all 30 models.  Thus reducing sampler contraction can
remove an apparent reversal benefit without changing the learned model or the
recorded direction, directly exposing the role of $B_\kappa$.
Score scaling under finite-step DDIM \cite{song2021ddim} is not exact global power tempering: the
Gaussian calibration retains nonzero shape residuals, especially for
anisotropic targets (Appendix~\ref{app:target-power-pilot}); we therefore do
not identify $\zeta$ with the global tempering parameter $\alpha$.

\section{Limitations and future work}
\label{sec:limitations}

The theory here deliberately studies power tempering as a particularly clean
idealization of mode-seeking inference. This restriction makes the
self-training direction exactly identifiable: under NLL training it is a
R\'enyi-entropy direction, and its geometry can therefore be analyzed without
conflating sampler effects with additional approximation error. Real inference
procedures need not be density reweightings, much less exact power temperings.
Extending the analysis to more general sampler maps---including finite-step
diffusion solvers \cite{song2021ddim,karras2022edm,lu2022dpmsolver},
guidance \cite{dhariwal2021guidance,hosalimans2022cfg}, truncation, and other
nonlinear inference transformations---is therefore an important next step.
Our local decomposition also holds the inference configuration $\kappa$
fixed while varying the reversal weight; jointly re-optimizing guidance,
resolution, temperature, or other sampler hyperparameters with the
reversal weight lies outside the present analysis.

A separate restriction is the training objective. The exact R\'enyi-entropy
identities are NLL statements, whereas diffusion models are typically trained
with denoising or score-matching objectives. This distinction does not remove
the basic obstruction: Appendix~\ref{app:score-gaussian} shows that the same
Gaussian mode-seeking counterexample persists for the population diffusion
score loss, for any nontrivial positive weighting over noise levels. Together
with the loss-agnostic compositional formulation of Section~\ref{sec:framework},
this suggests that substantial parts of the paper should admit score-based
analogues. Establishing the corresponding score-based geometry, correction
criteria is an important direction for future work.

At the same time, the simplicity of power tempering is part of the point. Even
in this unusually favorable setting, knowing that inference is mode-seeking
does not determine whether reversing self-training corrects the model: the
answer still depends on the model's position relative to the target, and
improvement of the emitted distribution can instead arise from sampler
compensation. Thus the central theme of the paper is that simple recipes such
as ``self-train and reverse'' are unlikely to admit a universal justification.
We view the present paper as the first step in a broader program studying what
information self-generated data reveal about the composed
model--sampler--training system. Future work will ask when this information is
sufficient to construct adaptive correction rules, how the relevant directions
change under realistic nonlinear samplers and score-based training objectives,
and whether multiple controlled inference perturbations can recover more than
the single direction exposed by one self-training experiment.

\section{Relation to prior work}
\label{sec:related}

NEON~\cite{alemohammad2025neon} introduced negative extrapolation from
self-training; our focus is the information contained in that displacement
and when its reversal reflects model correction rather than sampler
compensation. This differs from the broader literature on self-consuming
generative models, which primarily studies repeated retraining, collapse,
and stabilization across generations
\cite{alemohammad2024mad,shumailov2024collapse,bertrand2024stability,
gerstgrasser2024collapse,fu2024selfconsuming,gillman2024selfcorrecting,
kazdan2025collapse,zhu2026reflow}. Here a single short self-training step is
instead treated as a local probe of the current model--sampler--training
composition.

Our separation of represented and emitted distributions is also consistent
with diffusion models, where learning a score or denoising field and
numerically realizing the generative law are distinct operations
\cite{ho2020ddpm,song2021sde,song2021ddim,karras2022edm,
lu2022dpmsolver,chen2023sampling}. The theoretical analysis connects
R\'enyi entropy and tempering \cite{renyi1961,neal2001ais}, transformation
geometry \cite{papamakarios2021flows}, elliptical and Fisher--Rao geometry
\cite{fang1990elliptical,amari2016information,skovgaard1984,
calvo1991,pennec2006}, and high-dimensional covariance theory
\cite{ledoitwolf2004,marchenko1967,baisilverstein2010}. These tools are used
here to isolate when a self-training direction reflects model error and when
sample improvement instead comes from the inference procedure.

\section{Conclusion}
\label{sec:conclusion}

Mode-seeking inference alone does not determine whether reversing
self-training is beneficial: the sign depends on the relation between the
current model, the target, and the inference procedure. Under power
tempering, our analysis makes this dependence explicit and identifies
settings---including finite-data high-dimensional Gaussians---where reversal
genuinely corrects model error. At the same time, reversal can improve emitted
samples even while worsening the represented model, by compensating for
sampler distortion. A self-training displacement should therefore be viewed
as a measurement of the composed model--sampler--training system, rather than
as a universal estimate of model error.
\clearpage
\subsection*{AI use statement}

In this work, we used generative AI tools to assist with writing portions of the experimental code and verifying mathematical proofs.

Additionally, we used these tools to support literature discovery and analysis, identify overlapping work, and critically assess the significance of the proposed contributions. After the authors had established the initial theoretical and experimental claims and the paper's positioning, we used generative AI to prepare initial drafts of selected manuscript sections. We subsequently used AI-assisted feedback to explore possible directions for strengthening the contribution and to improve the manuscript's clarity and presentation.

We retained control over the research, evaluated the AI-generated suggestions, and made the final decisions about the claims, positioning, and content included in the paper. We take responsibility for the final content of this work, including all text, mathematical claims, proofs, code, experimental results, and other artifacts produced with the aid of generative AI.

\clearpage

\appendix

\section{Density reweighting and mode seeking}
\label{app:density-reweighting}

The main text focuses on power tempering because it gives the
R\'enyi-entropy direction used throughout the geometric analysis.
Here we record the more general structure behind that result.

\subsection{General density-reweighting samplers}
\label{app:density-reweighting-general}

Consider a sampler whose output depends pointwise on the current model
density:
\[
    q_\theta(x)
    =
    \frac{\omega(p_\theta(x))}{Z(\theta)},
    \qquad
    Z(\theta)
    =
    \int \omega(p_\theta(x))\,\dd x,
\]
for some profile $\omega:\mathbb R_+\to\mathbb R_+$.
We call such a sampler \emph{density reweighting}.

The potential associated with $\omega$ is obtained by choosing
$\Phi$ such that
\[
    \Phi'(t)=\frac{\omega(t)}{t},
\]
and defining
\[
    H_\Phi(p)
    :=
    -\int \Phi(p(x))\,\dd x.
\]
The appearance of $\omega(t)/t$ is forced by NLL training: the
sampler contributes the factor $\omega(p_\theta)$, while the score
satisfies
\[
    \nabla_\theta\log p_\theta
    =
    \frac{\nabla_\theta p_\theta}{p_\theta}.
\]

\begin{proposition}[Density reweighting induces a potential]
\label{prop:density-reweighting-potential}
Let $q_{\theta_r}$ be generated from the base model $p_{\theta_r}$
by a density-reweighting sampler with profile $\omega$. Suppose
$p_\theta$ is differentiable in $\theta$, that
\[
    \int
    \omega(p_\theta)
    \|\nabla_\theta\log p_\theta\|
    \,\dd x
    <\infty
\]
locally uniformly near $\theta_r$, and that
$\int\Phi(p_\theta)\dd x$ is finite. Then population NLL
self-training on samples from $q_{\theta_r}$ satisfies
\[
    \nabla\Rsyn(\theta_r)
    =
    \frac{1}{Z(\theta_r)}
    \nabla_\theta H_\Phi(p_\theta)
    \Big|_{\theta=\theta_r}.
\]
Thus the synthetic-data gradient is generated by the scalar
functional $H_\Phi$.
\end{proposition}

\begin{proof}
Write
\[
    u_\theta(x)
    :=
    \nabla_\theta\log p_\theta(x).
\]
Since
\[
    \Phi'(p_\theta)
    =
    \frac{\omega(p_\theta)}{p_\theta},
\]
differentiation under the integral gives
\[
    \nabla_\theta
    \int \Phi(p_\theta)\,\dd x
    =
    \int
    \frac{\omega(p_\theta)}{p_\theta}
    \nabla_\theta p_\theta
    \,\dd x
    =
    \int
    \omega(p_\theta)u_\theta
    \,\dd x.
\]
On the other hand, the synthetic distribution is held fixed at
$q_{\theta_r}$ during self-training, so
\[
    \nabla\Rsyn(\theta_r)
    =
    -\E_{q_{\theta_r}}[u_{\theta_r}]
    =
    -\frac{1}{Z(\theta_r)}
    \int
    \omega(p_{\theta_r})u_{\theta_r}
    \,\dd x.
\]
Since
\[
    H_\Phi(p_\theta)
    =
    -\int\Phi(p_\theta)\,\dd x,
\]
the claimed identity follows.
\end{proof}

Power tempering is the special case
\[
    \omega(t)=t^\alpha.
\]
Then one may take
\[
    \Phi(t)=\frac{t^\alpha}{\alpha},
    \qquad
    H_\Phi(p)
    =
    -\frac1\alpha\int p^\alpha\,\dd x
    =
    -\frac1\alpha
    \exp\!\bigl((1-\alpha)H_\alpha(p)\bigr).
\]
Differentiating this monotone transform recovers the main-text identity
\[
    \nabla\Rsyn(\theta_r)
    =
    \frac{\alpha-1}{\alpha}
    \nabla_\theta H_\alpha(p_\theta)
    \Big|_{\theta=\theta_r}.
\]

\subsection{Mode seeking and convexity of the potential}
\label{app:mode-seeking-potential}

Density reweighting also makes precise the sense in which the sampler
favors high-density regions. Indeed,
\[
    \frac{q_\theta(x)}{p_\theta(x)}
    =
    \frac{1}{Z(\theta)}
    \frac{\omega(p_\theta(x))}{p_\theta(x)}.
\]
Thus the relative weight assigned by the sampler increases with the
model density precisely when $t\mapsto\omega(t)/t$ is nondecreasing.

\begin{proposition}[Mode seeking and the associated potential]
\label{prop:mode-seeking-convexity}
For a density-reweighting sampler, the ratio $q_\theta/p_\theta$ is
a nondecreasing function of $p_\theta$ if and only if
\[
    \frac{\omega(t)}{t}
\]
is nondecreasing. Equivalently, $\Phi$ is convex, and hence
$H_\Phi$ is concave.
\end{proposition}

\begin{proof}
The normalization $Z(\theta)$ is independent of $x$, so monotonicity
of $q_\theta(x)/p_\theta(x)$ as a function of $p_\theta(x)$ is exactly
monotonicity of $\omega(t)/t$. Since
\[
    \Phi'(t)=\frac{\omega(t)}{t},
\]
this is equivalent to convexity of $\Phi$. The functional
$H_\Phi(p)=-\int\Phi(p)$ is therefore concave.
\end{proof}

This includes the broader monotone class
\[
    q(x)
    \propto
    f(\log p(x))\,p(x),
\]
with $f$ nondecreasing. In this case
\[
    \omega(t)=t f(\log t),
    \qquad
    \frac{\omega(t)}{t}=f(\log t),
\]
so the sampler is mode seeking in the above sense.

Power tempering corresponds to
$f(s)=e^{(\alpha-1)s}$. Hence $\alpha>1$ gives a mode-seeking
sampler, while $\alpha<1$ reverses the monotonicity.

\subsection{Mode seeking does not imply gradient anti-alignment}
\label{app:monotone-covariance-gap}

The preceding characterization concerns how the sampler reweights
points according to the \emph{current model density}. It does not imply
that this reweighting is monotone in the local error direction relative
to the target. This distinction is the source of the gap discussed in
Section~\ref{sec:gaussian-scale-example} and we make explicit here which part of the theoretical argument of
NEON~\cite{alemohammad2025neon} fails. Their Theorem~1 is a conditional
statement. In their notation, if the sampler bias forms an obtuse angle
with the model error,
\[
    \cos\phi<0,
\]
then, for a sufficiently accurate base model, the real- and
synthetic-data population gradients are anti-aligned:
\[
    s=\langle r_d,Pr_s\rangle<0.
\]

The additional step is NEON's Theorem~2, which claims that the required
angle condition follows from mode seeking itself. For the NLL
specialization, they consider
\[
    q(x)\propto
    f\!\left(\log p_{\theta_r}(x)\right)p_{\theta_r}(x),
    \qquad f \ \text{nondecreasing},
\]
write $\theta_r=\theta_\star+\varepsilon$, and define
\[
    B(x)=
    \varepsilon^\top
    u_{\theta_\star}(x),
    \qquad
    u_{\theta_\star}(x)
    =
    \nabla_\theta\log p_\theta(x)\big|_{\theta_\star}.
\]
The local expansion is
\[
    \log p_{\theta_r}(x)
    =
    \log p_{\theta_\star}(x)
    +B(x)
    +O(\|\varepsilon\|^2).
\]
The proof then treats
\[
    f\!\left(\log p_{\theta_r}(x)\right)
\]
as a nondecreasing function of the scalar $B(x)$ and invokes a monotone
covariance inequality.

This is the gap. Monotonicity of $f$ in the \emph{current log-density}
does not imply monotonicity in the \emph{local error tangent} $B$.
The term $\log p_{\theta_\star}(x)$ is generally nonconstant and of
order one, so two points can be ordered oppositely by
$\log p_{\theta_r}(x)$ and by $B(x)$. Consequently the covariance sign
used in the proof of Theorem~2 is not determined by mode seeking alone.

The failure already occurs for global power tempering. Let
\[
    p_t=\mathcal N(0,e^{2t}\Sigma_\star),
    \qquad
    p_\star=p_0,
\]
and take a slightly over-dispersed model $t_r=\varepsilon>0$. Put
\[
    r^2(x)=x^\top\Sigma_\star^{-1}x.
\]
At the target,
\[
    \frac{\partial}{\partial t}\log p_t(x)\Big|_{t=0}
    =
    r^2(x)-d,
\]
so
\[
    B(x)=\varepsilon\bigl(r^2(x)-d\bigr)
\]
is strictly increasing in $r^2$. On the other hand, for
\[
    q_\alpha(x)
    \propto
    p_\varepsilon(x)^\alpha,
    \qquad \alpha>1,
\]
the sampler weight relative to the current model is
\[
    \frac{q_\alpha(x)}{p_\varepsilon(x)}
    =
    C\exp\!\left(
       -\frac{\alpha-1}{2e^{2\varepsilon}}\,r^2(x)
    \right),
\]
which is strictly \emph{decreasing} in $r^2$, and hence decreasing in
$B$. In particular,
\[
    \operatorname{Cov}_{p_\star}
    \!\left(
        \frac{q_\alpha}{p_\varepsilon},B
    \right)<0,
\]
the opposite of the covariance sign used in the proof of NEON's
Theorem~2.

The contradiction can also be seen directly at the level of the two
population gradients. Along this one-dimensional scale family,
\[
    R'_{\rm data}(\varepsilon)
    =
    d\bigl(1-e^{-2\varepsilon}\bigr)>0,
\]
while NLL training on the power-tempered law gives
\[
    R'_{\rm syn}(\varepsilon)
    =
    d\left(1-\frac1\alpha\right)>0.
\]
Thus the real- and synthetic-data gradients have the \emph{same} sign,
so for any positive scalar preconditioner
\[
    \langle r_d,Pr_s\rangle>0.
\]
Self-training moves the over-dispersed model toward the target, whereas
negative extrapolation moves it farther away.

Power tempering with $\alpha>1$ satisfies exactly the monotone
mode-seeking hypothesis of NEON's Theorem~2, and $\varepsilon$ may be
arbitrarily small. Hence Theorem~2 is false as stated. Since the
mode-seeking guarantee for NEON is obtained by combining Theorem~2
with the conditional anti-alignment result of Theorem~1, that combined
guarantee does not follow. This does not invalidate the empirical
improvements reported by NEON; it shows that mode seeking alone cannot
provide their claimed universal sign argument.

\subsection{Score-based Gaussian analogue}
\label{app:score-gaussian}

The counterexample above was stated for NLL training.  We now show that its
sign does not depend on that choice of objective.  Score matching fits the
spatial score $\nabla_x\log p(x)$ rather than the log likelihood
\cite{hyvarinen2005score}; denoising score matching gives an equivalent
population objective for perturbed data and underlies modern diffusion
training \cite{vincent2011denoising,ho2020ddpm,song2021sde}.

Let the forward perturbation at noise level $\tau$ be
\[
    X_\tau=m_\tau X_0+\sigma_\tau Z,
    \qquad Z\sim\mathcal N(0,1),
\]
and write $r_\tau$ for the resulting marginal when $X_0\sim r$.  Its
population score is
\[
    s_r(x,\tau):=\partial_x\log r_\tau(x).
\]
For a model score $s_u$, consider the weighted population score loss
\[
    \mathcal L_{\rm sc}(u;r)
    :=
    \frac12\int_0^T
    \lambda(\tau)\,
    \E_{r_\tau}
    \!\left[
        \bigl(s_u(X,\tau)-s_r(X,\tau)\bigr)^2
    \right]
    \dd\tau,
\]
where $\lambda(\tau)\ge0$ is nontrivial.  The usual denoising formulation
differs from this population objective only by terms independent of the
model parameters, so it has the same population training gradient.

Return now to the Gaussian scale family
\[
    p_t=\mathcal N(0,e^{2t}),
    \qquad
    p_\star=\mathcal N(0,1),
\]
and write $v=e^{2t}$.  If the candidate model has scale parameter $u$, its
perturbed variance and exact score are
\[
    c_\tau(e^{2u})
    :=
    m_\tau^2e^{2u}+\sigma_\tau^2,
    \qquad
    s_u(x,\tau)
    =
    -\frac{x}{c_\tau(e^{2u})}.
\]
Direct differentiation at the base model $u=t$ gives
\[
    \frac{\partial}{\partial u}
    \mathcal L_{\rm sc}(u;p_\star)
    \Big|_{u=t}
    =
    2v(v-1)
    \int_0^T
    \lambda(\tau)
    \frac{m_\tau^4}{c_\tau(v)^3}
    \dd\tau.
\]
Hence the population score-risk derivative has the sign of $t$.

Power tempering the same base model gives
\[
    q_\alpha
    =
    \mathcal N\!\left(0,\frac{v}{\alpha}\right),
    \qquad \alpha>1.
\]
Holding this synthetic distribution fixed during self-training,
\[
    \frac{\partial}{\partial u}
    \mathcal L_{\rm sc}(u;q_\alpha)
    \Big|_{u=t}
    =
    2v^2\left(1-\frac1\alpha\right)
    \int_0^T
    \lambda(\tau)
    \frac{m_\tau^4}{c_\tau(v)^3}
    \dd\tau
    >0.
\]
Thus score-based self-training again contracts the model scale and reversal
expands it.  Consequently,
\[
    t<0:\ \text{reversal improves the population score loss},
    \qquad
    t>0:\ \text{reversal worsens it}.
\]
The same mode-seeking sampler therefore produces opposite reversal signs
under diffusion score matching just as under NLL.  The failure of
mode seeking to determine the corrective direction is not an artifact of
likelihood training.

\section{Transformation, scale, and elliptical-family results}
\label{app:transformation-families}

\subsection{Entropy along a transformation family}

\begin{proposition}[Entropy along a transformation family]
\label{prop:transform-entropy}
Let $p_t=(T_t)_\#p_0$ be a smooth family with velocity field $v_t$, so that
$\partial_t p_t=-\nabla\!\cdot(p_tv_t)$. Under the decay conditions needed to
remove the boundary term,
\[
    \frac{d}{dt}H_\alpha(p_t)
      =\E_{q_{\alpha,t}}[\nabla\!\cdot v_t],
    \qquad q_{\alpha,t}\propto p_t^\alpha.
\]
\end{proposition}

\begin{proof}
Let $Z_\alpha(t)=\int p_t^\alpha$. Using $\partial_t p_t=-\nabla\cdot(p_tv_t)$ and integration by parts,
\[
\begin{aligned}
\dot Z_\alpha
&=-\alpha\int p_t^{\alpha-1}\nabla\cdot(p_tv_t)\\
&=\alpha(\alpha-1)\int p_t^{\alpha-1}\nabla p_t\cdot v_t\\
&=-(\alpha-1)\int p_t^\alpha\nabla\cdot v_t.
\end{aligned}
\]
Therefore
\[
\frac{d}{dt}H_\alpha(p_t)
=\frac{1}{1-\alpha}\frac{\dot Z_\alpha}{Z_\alpha}
=\frac{\int p_t^\alpha\nabla\cdot v_t}{\int p_t^\alpha}
=\E_{q_{\alpha,t}}[\nabla\cdot v_t].
\]
The decay assumptions are those needed to discard the boundary term.
\end{proof}

\subsection{Affine and scale families}
For $p_{\mu,A}(x)=|\det A|^{-1}g(A^{-1}(x-\mu))$,
\[
\int p_{\mu,A}(x)^\alpha\,dx
=|\det A|^{1-\alpha}\int g(y)^\alpha\,dy,
\]
so $H_\alpha(p_{\mu,A})=H_\alpha(g)+\log|\det A|$. The differential in an affine direction $U$ is $D_AH_\alpha[U]=\tr(A^{-1}U)$; it vanishes on the tangent space of determinant-preserving deformations.

For the isotropic scale family $p_t(x)=e^{-dt}g(e^{-t}x)$, this reduces to $H_\alpha(p_t)=H_\alpha(g)+dt$.

\subsection{Scale-family sign}

\begin{proposition}[Scale-family sign]
\label{prop:scale-family}
Suppose the truth and model belong to the same regular scale family, with true
scale $t_\star$. Then locally
\[
    \operatorname{sign} R_{\mathrm{data}}'(t)
      =\operatorname{sign}(t-t_\star).
\]
Consequently entropy-increasing reversal helps a sufficiently close model
exactly when it is too narrow. If $g\propto e^{-V}$ with differentiable convex
$V$ minimized at the origin, the same sign relation holds globally, strictly
away from $t_\star$ under the usual nondegeneracy conditions.
\end{proposition}

\begin{proof}
By re-centering the scale parameter, take $t_\star=0$. Regular likelihood theory gives $\Rdata'(0)=0$ and $\Rdata''(0)=I_{\rm scale}>0$, so $\Rdata'(t)=I_{\rm scale}t+o(t)$, proving the local sign statement.

For the global log-concave statement write $g(x)=Z^{-1}e^{-V(x)}$, where $V$ is differentiable, convex, and minimized at the origin. With $X\sim g$,
\[
\Rdata(t)=dt+\E[V(e^{-t}X)]+C,
\qquad
\Rdata'(t)=d-\E[e^{-t}X\cdot\nabla V(e^{-t}X)].
\]
Set $s=e^{-t}$ and $K_x(s)=s\,x\cdot\nabla V(sx)$. Convexity and minimality at $0$ give $x\cdot\nabla V(sx)\ge0$, and where the Hessian exists,
\[
K_x'(s)=x\cdot\nabla V(sx)+s\,x^\top\nabla^2V(sx)x\ge0.
\]
The same monotonicity follows by approximation for differentiable convex $V$. Integration by parts under $g$ gives $\E[X\cdot\nabla V(X)]=d$, hence $\E K_X(1)=d$. Therefore $t>0$ implies $s<1$ and $\Rdata'(t)\ge0$, while $t<0$ implies $\Rdata'(t)\le0$. Strictness follows unless the scale direction is degenerate.
\end{proof}

\subsection{Elliptical radial-score calculation}
\label{app:elliptical-proof}
\begin{proof}[Proof of Proposition~\ref{prop:elliptical-risk}]
For $u_\Sigma(x)=x^\top\Sigma^{-1}x$ and $\Sigma(a)=e^a\Sigma$,
\[
-\log p_{\Sigma(a)}(x)
=C+\frac12\log\det\Sigma+\frac{ad}{2}-\log f(e^{-a}u_\Sigma(x)).
\]
Differentiating at $a=0$ yields
\[
\frac d2+u_\Sigma(x)\frac{f'(u_\Sigma(x))}{f(u_\Sigma(x))}
=\frac d2-\psi_f(u_\Sigma(x)).
\]
Taking expectation under $p_\star$ gives the stated criterion. At correct specification, differentiating the population cross-entropy at its minimum gives $\E_{p_\Sigma}\psi_f(u_\Sigma)=d/2$.
\end{proof}

\section{Gaussian information geometry}
\label{app:spd}

The space $\SPD(d)$ is an open cone in the vector space of symmetric matrices, so $T_\Sigma\SPD(d)$ is the space of symmetric matrices. The Fisher metric for centered Gaussians is $g_\Sigma(U,V)=\tfrac12\tr(\Sigma^{-1}U\Sigma^{-1}V)$. Its Riemannian exponential and logarithm are
\begin{align}
    \operatorname{Exp}_\Sigma(U)
    &=\Sigma^{1/2}\exp(\Sigma^{-1/2}U\Sigma^{-1/2})\Sigma^{1/2},\\
    \operatorname{Log}_\Sigma(P)
    &=\Sigma^{1/2}\log(\Sigma^{-1/2}P\Sigma^{-1/2})\Sigma^{1/2},
\end{align}
and the squared distance is
\begin{equation}
    d_F^2(\Sigma,P)
    =\frac12\left\|\log(\Sigma^{-1/2}P\Sigma^{-1/2})\right\|_F^2.
\end{equation}
These are standard formulas for the affine-invariant metric \cite{skovgaard1984,calvo1991,pennec2006}.

\subsection{Fisher Rao criteria for Gaussian}
\label{app:gaussian-fisher-sign}
\begin{proposition}[Fisher--Rao alignment]
\label{prop:det-criterion}
For centered Gaussian model and target,
\[
    g_\Sigma(\mathrm{Log}_\Sigma\Sigstar,\Sigma)
    =\frac12\log\frac{\det\Sigstar}{\det\Sigma},
\]
so isotropic expansion points toward the target in Fisher--Rao geometry iff
\[
    \det\Sigma<\det\Sigstar .
\]
\end{proposition}

\begin{proof}
Along the isotropic curve
\[
    \Sigma(a)=e^a\Sigma,
\]
the tangent at $a=0$ is $\Sigma$. Let
\[
    X=\Log_\Sigma(\Sigstar).
\]
Then
\[
    g_\Sigma(X,\Sigma)
    =\frac12\tr(\Sigma^{-1}X)
    =\frac12\tr\log(\Sigma^{-1/2}\Sigstar\Sigma^{-1/2})
    =\frac12\log\frac{\det\Sigstar}{\det\Sigma}.
\]
\end{proof}

\section{Finite-sample and high-dimensional Gaussian learning}
\label{app:wishart}

All asymptotic calculations can be performed after whitening, so take $\Sigstar=I$. For fixed zero mean and $X_i\sim\Normal(0,I)$,
\begin{equation}
    nS_n=\sum_{i=1}^nX_iX_i^\top\sim W_d(I,n).
\end{equation}
For the two-dimensional non-whitened target used in the experiments, $\det\Sigstar=3$ and the Wishart determinant moment gives
\begin{equation}
    \E[\det\whSigma]=3\frac{n-1}{n}.
\end{equation}

\subsection{Proof of Proposition~\ref{prop:fixed-d-wishart}}

Write $S_n=I+E_n$. For fixed $d$, $\|E_n\|=O_p(n^{-1/2})$. Taylor expansion gives
\begin{equation}
    \log\det(I+E_n)=\tr E_n+O_p(n^{-1}),
\end{equation}
and
\begin{equation}
    \tr(I+E_n)^{-1}-d=-\tr E_n+O_p(n^{-1}).
\end{equation}
Moreover,
\begin{equation}
    \tr S_n=\frac1n\sum_{i=1}^n\|X_i\|^2,
\end{equation}
so $n\tr S_n\sim\chi^2_{nd}$ and hence
\begin{equation}
    \sqrt n\,\tr E_n
    =\sqrt n(\tr S_n-d)
    \Rightarrow\Normal(0,2d).
\end{equation}
Slutsky's theorem yields \eqref{eq:fixed-d-clt}; symmetry of the limiting normal gives probability $1/2$ for either favorable sign.

\subsection{Two-dimensional Monte Carlo}\label{app:wishart-2d}

The finite-dimensional phenomenon is visible at practical sample sizes. For
\begin{equation}
    \Sigstar=\begin{bmatrix}2&1\\1&2\end{bmatrix},
\end{equation}
$2000$ independent datasets at each sample size gave
\begin{table}[H]
\centering
\caption{Finite-sample centered Gaussian MLE. The Fisher-alignment sign remains two-sided as the estimator concentrates, consistent with Proposition~\ref{prop:fixed-d-wishart}.}
\label{tab:wishart-main}
\small
\begin{tabular}{rcc}
\toprule
$n$ & $\det\whSigma$ (mean $\pm$ sd) & $\Pr(\det\whSigma<3)$\\
\midrule
500   & $2.9878\pm0.2647$ & 53.4\%\\
1000  & $2.9995\pm0.1912$ & 50.4\%\\
5000  & $3.0006\pm0.0829$ & 49.8\%\\
10000 & $2.9994\pm0.0599$ & 50.7\%\\
\bottomrule
\end{tabular}
\end{table}

\subsection{Proof of Theorem~\ref{thm:mp-unsmoothed}}

Under $d/n\to\gamma\in(0,1)$, the empirical spectral distribution of $S_n$ converges almost surely to the Marchenko--Pastur law $\mu_{\mathrm{MP},\gamma}$, whose support is
\begin{equation}
    \bigl[(1-\sqrt\gamma)^2,(1+\sqrt\gamma)^2\bigr].
\end{equation}
For Gaussian sample covariance matrices the extreme eigenvalues converge to the support edges, so the functions $\lambda\mapsto\lambda^{-1}$ and $\lambda\mapsto\log\lambda$ are uniformly controlled for $\gamma<1$. Therefore
\begin{equation}
    \frac1d\tr(S_n^{-1})
    \longrightarrow
    \int\lambda^{-1}\,\dd\mu_{\mathrm{MP},\gamma}(\lambda)
    =\frac1{1-\gamma},
\end{equation}
and
\begin{equation}
    \frac1d\log\det S_n
    \longrightarrow
    \int\log\lambda\,\dd\mu_{\mathrm{MP},\gamma}(\lambda)
    =\frac{1-\gamma}{\gamma}\log\frac1{1-\gamma}-1.
\end{equation}
The second expression is strictly negative for $\gamma\in(0,1)$. This proves Theorem~\ref{thm:mp-unsmoothed}.
\subsection{Proof of Remark~\ref{rem:mp-smoothing}}

Set $t=h^2>0$. Since the target has been whitened to $I$, the Gaussian
NLL criterion of Corollary~\ref{cor:trace} says that isotropic
expansion of
\[
    \Sigma_h=S_n+tI
\]
is favorable exactly when
\[
    \frac1d\tr(S_n+tI)^{-1}>1.
\]
Under $d/n\to\gamma\in(0,1)$, Marchenko--Pastur convergence gives
\[
    \frac1d\tr(S_n+tI)^{-1}
    \longrightarrow
    m_\gamma(t)
    :=
    \int\frac{1}{\lambda+t}\,
    \dd\mu_{\mathrm{MP},\gamma}(\lambda),
\]
where
\[
    m_\gamma(t)
    =
    \frac{
        \sqrt{(t+1+\gamma)^2-4\gamma}
        -(t+1-\gamma)
    }{2\gamma t}.
\]
Moreover,
\[
    m_\gamma'(t)
    =
    -\int\frac{1}{(\lambda+t)^2}\,
    \dd\mu_{\mathrm{MP},\gamma}(\lambda)
    <0,
\]
so there is at most one transition. Solving $m_\gamma(t)=1$ gives
\[
    t_\star=\frac{\gamma}{1+\gamma}.
\]
Hence
\[
    m_\gamma(t)>1
    \iff
    t<\frac{\gamma}{1+\gamma},
\]
which proves the stated phase boundary.

\section{Distribution-space setup, training response, and sampler Jacobian}
\label{app:operators}

We formalize the additive tangent convention used in
Section~\ref{sec:framework}.
Let $\PP$ denote a regular class of strictly positive densities with respect
to $\mu$. Since $\int p\,d\mu=1$, its ambient tangent space is
\[
    T_p\PP\simeq\V
    =
    \left\{h:\int h\,d\mu=0\right\}
\]
at every $p$. When $h/p\in L^2(p)$, the Fisher metric is
\[
    g_p^F(h,k)=\int \frac{hk}{p}\,d\mu.
\]
The map $h\mapsto h/p$ identifies this additive tangent with the
relative-density representation $L_0^2(p)$; the two are coordinates for the
same infinitesimal perturbation, not different residuals.

Let $G:\Theta\to\PP$ be the model map, $G(\theta)=p_\theta$. Then
\[
    T_p\mathcal M
    =
    \range(DG_{\theta_r})
    \subset\V.
\]
Let $\mathcal T_p(q)$ be the population endpoint of the prescribed short
training procedure, initialized at $p$ and trained on $q$. After centering any
baseline drift so that $\mathcal T_p(p)=p$, define
\[
    A
    :=
    D\mathcal T_p|_p
    :
    \V\to T_p\mathcal M.
\]
This definition is loss- and optimizer-agnostic.

For NLL, the Fisher projection arises explicitly. Write the model density
tangents as
\[
    h_i
    =
    \partial_{\theta_i}p_\theta|_{\theta_r}
    =
    p\,u_i,
    \qquad
    u_i
    =
    \partial_{\theta_i}\log p_\theta|_{\theta_r}.
\]
If the training law is
\[
    q=p+\epsilon h,
    \qquad
    h\in\V,
\]
then at $\theta_r$
\[
    \partial_{\theta_i}\E_q[-\log p_\theta]
      =
      -\epsilon\int h\,u_i\,d\mu
      =
      -\epsilon g_p^F(h,h_i).
\]
Linearizing the score equations after a parameter displacement
$\Delta\theta$ gives
\[
    \sum_j
    g_p^F(h_i,h_j)\Delta\theta_j
      =
      \epsilon g_p^F(h,h_i).
\]
In the infinitesimal exact-refit limit, the induced density response is
the Fisher projection $\Pi_p^F h$; an appropriately scaled natural-gradient
step gives the corresponding short-step response
$\rho\Pi_p^F h$.  Finite Euclidean, SGD, or Adam training generally
depends on the optimizer and parametrization and need not be proportional
to this projection.  We therefore retain $A$ as the empirical linear
training-response operator in the general theory.

For the self-training setting, write
\[
    \delta:=p-p_\star,
    \qquad
    \varphi:=S_\kappa(p)-p.
\]
Thus the residual of the emitted distribution is
\[
    S_\kappa(p)-p_\star
    =
    \delta+\varphi.
\]

The sampler derivative is
\[
    B_\kappa
    :=
    DS_\kappa|_p
    :
    T_p\PP
    \to
    T_{S_\kappa(p)}\PP.
\]
Using the affine identification of both tangent spaces with $\V$, we write
simply $B_\kappa:\V\to\V$. If
\[
    S_\kappa(p)
    =
    p+\varphi_\kappa(p),
\]
then
\[
    B_\kappa
    =
    I+D\varphi_\kappa(p).
\]
Consequently the reversal displacement $-wA\varphi$ is seen after sampling as
\[
    -wB_\kappa A\varphi
\]
to first order. Writing
\[
    p_{\rev}(w)
    :=
    p-wA\varphi
\]
at this order gives the bookkeeping summarized in
Table~\ref{tab:composition}.

\begin{table}[t]
\centering
\caption{First-order decomposition around the base model
$p=p_{\theta_r}$.}
\label{tab:composition}
\footnotesize
\setlength{\tabcolsep}{2pt}

\begin{tabular}{
@{}
>{\raggedright\arraybackslash}p{0.40\columnwidth}
>{\raggedright\arraybackslash}p{0.54\columnwidth}
@{}
}
\toprule
Stage & Residual relative to $p_\star$ \\
\midrule

Target
& $0$ \\

Base model
& $\delta$ \\

Deployed sampler
& $\delta+\varphi$ \\

Self-trained model
& $\delta+A\varphi$ \\

Reversed model
& $\delta-wA\varphi$ \\

Deployed reversed model
& $\delta+\varphi-wB_\kappa A\varphi$ \\

\bottomrule
\end{tabular}
\end{table}

The last two rows make explicit the distinction used in the main text:
reversal changes the represented model by $-wA\varphi$, whereas the change
visible after applying the inference procedure is
$-wB_\kappa A\varphi$.

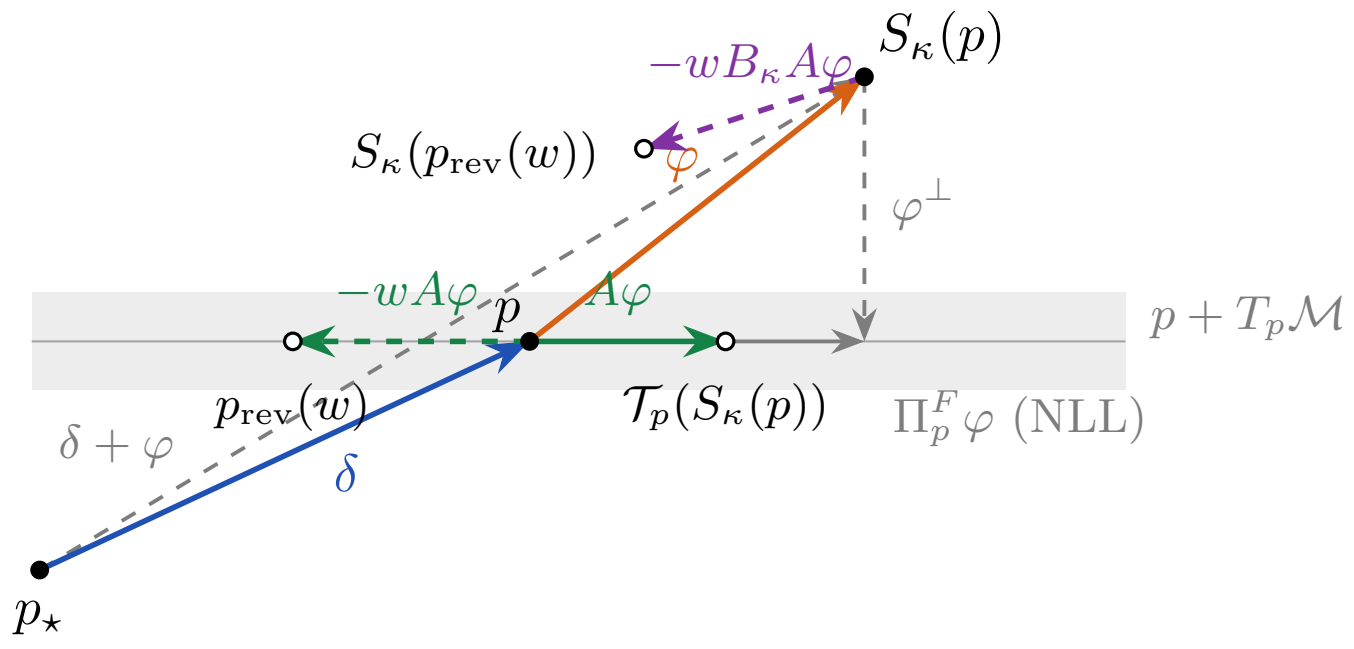
\begin{figure}[H]
\centering
\begin{tikzpicture}[scale=1.08]
  \begin{scope}[on background layer]
    \fill[black!7] (-3.05,-0.30) rectangle (3.65,0.30);
  \end{scope}
  \draw[black!35] (-3.05,0) -- (3.65,0);
  \node[slab,cgy,anchor=west] at (3.68,0.14)
      {$p+T_p\mathcal M$};

  \coordinate (pstar) at (-3.0,-1.40);
  \coordinate (p)     at (0,0);
  \coordinate (q)     at (2.05,1.62);
  \coordinate (Pphi)  at (2.05,0);
  \coordinate (ps)    at (1.20,0);
  \coordinate (prev)  at (-1.45,0);
  \coordinate (qrev)  at (0.70,1.18);

  \draw[tvec,cgy,dashed] (pstar) -- (q)
     node[pos=.18,above left=-2pt,slab,cgy]
     {$\delta+\varphi$};

  \draw[vec,cdelta] (pstar) -- (p)
     node[pos=.57,below right=-2pt,lab,cdelta]
     {$\delta$};
  \draw[vec,cphi] (p) -- (q)
     node[pos=.55,above left=-2pt,lab,cphi]
     {$\varphi$};

  \draw[tvec,cgy,dashed] (q) -- (Pphi)
     node[pos=.48,right=1pt,slab,cgy]
     {$\varphi^\perp$};
  \draw[tvec,cgy] (p) -- (Pphi);
  \node[slab,cgy,anchor=west] at (2.10,-0.47)
     {$\Pi_p^F\varphi$ (NLL)};

  \draw[vec,cA] (p) -- (ps)
     node[pos=.45,above=1pt,slab,cA]
     {$A\varphi$};
  \draw[vec,cA,dashed] (p) -- (prev)
     node[pos=.52,above=1pt,slab,cA]
     {$-wA\varphi$};
  \draw[vec,cB,dashed] (q) -- (qrev)
     node[pos=.52,above=1pt,slab,cB]
     {$-wB_\kappa A\varphi$};

  \node[dot] at (pstar) {};
  \node[lab,below=2pt] at (pstar) {$p_\star$};

  \node[dot] at (p) {};
  \node[lab,above left=-3pt] at (p) {$p$};

  \node[dot] at (q) {};
  \node[lab,above right=-2pt] at (q) {$S_\kappa(p)$};

  \node[odot] at (ps) {};
  \node[slab,below=4pt] at (ps)
      {$\mathcal T_p(S_\kappa(p))$};

  \node[odot] at (prev) {};
  \node[slab,below=4pt] at (prev)
      {$p_{\mathrm{rev}}(w)$};

  \node[odot] at (qrev) {};
  \node[slab,anchor=east] at (0.55,1.14)
      {$S_\kappa(p_{\mathrm{rev}}(w))$};

  \useasboundingbox (-4.05,-1.92) rectangle (4.25,2.18);
\end{tikzpicture}
\caption{Geometric view of Table~\ref{tab:composition}.
The sampler creates the displacement \textcolor{cphi}{$\varphi$}, while
self-training can record only the model-representable response
\textcolor{cA}{$A\varphi$}. For idealized NLL refitting, the grey construction
shows the Fisher projection $\Pi_p^F\varphi$. Reversal changes the model by
\textcolor{cA}{$-wA\varphi$}, whereas after inference the corresponding change
is \textcolor{cB}{$-wB_\kappa A\varphi$}.}
\label{fig:tangent-decomposition}
\end{figure}

Figure~\ref{fig:tangent-decomposition} also makes clear why the self-training
direction is not, in general, an estimate of the model residual $\delta$.
The sampler creates $\varphi$, training records its representable response
$A\varphi$, and inference acts again through $B_\kappa$.

Two limitations are explicit in this formulation. First,
\[
    \range(A)\subseteq T_p\mathcal M,
\]
so sampler error outside the locally representable model tangent cannot be
recorded directly in the weights. Second, $B_\kappa A\varphi$ need not remain
in $T_p\mathcal M$; the directions reachable in emitted-distribution space are
governed by the composed response
\[
    B_\kappa\circ A.
\]

\section{General directional criteria for the composed pipeline}
\label{app:two-risk-proofs}

This appendix gives the formal directional-derivative statements used in
Section~\ref{sec:framework}.  We write
\[
    \mathrm d\mathcal F_r[h]
\]
for the Fr\'echet differential of a functional $\mathcal F$ at the
distribution $r$, evaluated in the tangent direction $h$.

\subsection{General directional criterion}

Let
\[
    \mathcal R:\mathcal P\to\mathbb R
\]
be a differentiable model-risk functional and let
\[
    \mathcal J:\mathcal P\to\mathbb R
\]
be a differentiable functional used to evaluate emitted distributions.
For the sample discrepancy $D$ used in the main text,
\[
    \mathcal J(r):=D(p_\star,r).
\]

Recall from Appendix~\ref{app:operators} that, with
$a=A\varphi$ and $b=B_\kappa a$, the local reversal path satisfies
\[
    p_{\rm rev}(w)=p-wa+o(w),
\]
and
\[
    S_\kappa(p_{\rm rev}(w))
    =
    q-wb+o(w),
    \qquad
    q=S_\kappa(p).
\]

\begin{proposition}[General directional criteria]
\label{prop:general-directional-criteria}
Under the differentiability assumptions above,
\[
    \frac{\dd}{\dd w}
    \mathcal R(p_{\rm rev}(w))
    \Big|_{w=0}
    =
    -\,\mathrm d\mathcal R_p[a],
\]
and
\[
    \frac{\dd}{\dd w}
    \mathcal J\!\left(S_\kappa(p_{\rm rev}(w))\right)
    \Big|_{w=0}
    =
    -\,\mathrm d\mathcal J_q[b].
\]
Consequently, for sufficiently small positive reversal weights,
\[
    \text{model improves}
    \quad\Longleftrightarrow\quad
    \mathrm d\mathcal R_p[a]>0,
\]
and
\[
    \text{samples improve}
    \quad\Longleftrightarrow\quad
    \mathrm d\mathcal J_q[b]>0.
\]
\end{proposition}

\begin{proof}
The first-order expansions above and the Fr\'echet chain rule give
\[
    \mathcal R(p_{\rm rev}(w))
    =
    \mathcal R(p)
    -
    w\,\mathrm d\mathcal R_p[a]
    +
    o(w),
\]
and
\[
    \mathcal J\!\left(S_\kappa(p_{\rm rev}(w))\right)
    =
    \mathcal J(q)
    -
    w\,\mathrm d\mathcal J_q[b]
    +
    o(w).
\]
Differentiating at $w=0$ gives the stated identities.
\end{proof}

\subsection{Exact NLL specialization}

For population NLL, define
\[
    \mathcal R(r)
    :=
    \E_{p_\star}[-\log r(X)].
\]
For an additive density tangent $a$ at $p$,
\[
    \mathrm d\mathcal R_p[a]
    =
    -\int p_\star\,\frac{a}{p}\,\dd\mu.
\]
Writing $\delta=p-p_\star$ and using
$\int a\,\dd\mu=0$ gives
\begin{align*}
    \mathrm d\mathcal R_p[a]
    &=
    -\int (p-\delta)\frac{a}{p}\,\dd\mu\\
    &=
    \int\frac{\delta a}{p}\,\dd\mu\\
    &=
    g_p^F(\delta,a),
\end{align*}
where
\[
    g_p^F(h,k)
    =
    \int\frac{hk}{p}\,\dd\mu
\]
is the Fisher metric on additive density tangents defined in
Appendix~\ref{app:operators}.

Thus, for NLL,
\[
    \text{model improves}
    \quad\Longleftrightarrow\quad
    g_p^F(\delta,a)>0,
\]
which is the criterion used in Section~\ref{sec:framework}.

\subsection{Local quadratic geometry of the sample objective}

For the sample-side functional $\mathcal J$, assume that $\mathcal J$ is
twice Fr\'echet differentiable near $p_\star$, that
\[
    \mathrm d\mathcal J_{p_\star}=0,
\]
and that its second differential is continuous.  Write
\[
    e:=q-p_\star=\delta+\varphi .
\]
A first-order expansion of the differential around the target gives
\[
    \mathrm d\mathcal J_{p_\star+e}[b]
    =
    \mathrm d^2\mathcal J_{p_\star}[e,b]
    +
    o(\|e\|\,\|b\|).
\]
Define the local bilinear form
\[
    G_J(h,k)
    :=
    \mathrm d^2\mathcal J_{p_\star}[h,k].
\]
Since $q=p_\star+e$,
\[
    \mathrm d\mathcal J_q[b]
    =
    G_J(\delta+\varphi,b)
    +
    o(\|\delta+\varphi\|\,\|b\|).
\]
Hence, to leading order,
\[
    \text{samples improve}
    \quad\Longleftrightarrow\quad
    G_J(\delta+\varphi,b)>0.
\]

If $\mathcal J$ is exactly quadratic in a fixed linear embedding, the
remainder vanishes and this identity is exact.

\subsection{General density-reweighting entropy bridge}
\label{app:entropy-sampler-dominance}

We give here the general density-reweighting form underlying the
power-tempering result of Section~\ref{sec:entropy-sampler-dominance}.
The notation follows Appendix~\ref{app:density-reweighting}.

Let
\[
    q=S(p)=\frac{\omega(p)}{Z}
          =\frac{pL}{Z},
    \qquad
    L:=\frac{\omega(p)}{p}=\Phi'(p),
    \qquad
    Z=\E_p L,
\]
and write
\[
    \delta:=p-p_\star,
    \qquad
    \varphi:=q-p.
\]
Recall from Appendix~\ref{app:density-reweighting} that the associated
potential is
\[
    H_\Phi(p):=-\int \Phi(p(x))\,\dd x,
    \qquad
    \Phi'(t)=\frac{\omega(t)}{t}.
\]

\begin{proposition}[Density-reweighting entropy--Fisher bridge]
\label{prop:general-entropy-fisher-bridge}
Whenever the displayed quantities are finite,
\[
    Z\,g_p^F(\delta,\varphi)
    =
    \E_pL-\E_{p_\star}L
    =
    H_\Phi(p_\star)-H_\Phi(p)
    +D_\Phi(p_\star\Vert p),
\]
where
\[
    D_\Phi(p_\star\Vert p)
    :=
    \int
    \Bigl[
        \Phi(p_\star)-\Phi(p)
        -\Phi'(p)(p_\star-p)
    \Bigr]\dd x .
\]

If, in addition, $\varphi\in T_p\mathcal M$ and the local NLL training
response is
\[
    A=\rho\Pi_p^F,
    \qquad
    \rho>0,
\]
then $a=A\varphi=\rho\varphi$, and negative extrapolation improves the
model NLL exactly when
\[
    \E_{p_\star}L<\E_pL.
\]

For a mode-seeking density reweighting, $\Phi$ is convex by
Proposition~\ref{prop:mode-seeking-convexity}, so
$D_\Phi(p_\star\Vert p)\ge0$. Consequently,
\[
    H_\Phi(p)<H_\Phi(p_\star)
\]
is a sufficient, though not necessary, condition for model correction
under the above representability assumption.
\end{proposition}

\begin{proof}
Since
\[
    \frac{q}{p}=\frac{L}{Z}
\]
and $\int\delta\,\dd x=0$,
\begin{align*}
    Z\,g_p^F(\delta,\varphi)
    &=
    Z\int
       \frac{\delta\varphi}{p}\,\dd x\\
    &=
    Z\int
       \delta
       \left(\frac{q}{p}-1\right)\dd x\\
    &=
    \int (p-p_\star)L\,\dd x\\
    &=
    \E_pL-\E_{p_\star}L.
\end{align*}
Adding and subtracting
\[
    \int\bigl[\Phi(p_\star)-\Phi(p)\bigr]\dd x
\]
gives
\begin{align*}
    \E_pL-\E_{p_\star}L
    &=
    H_\Phi(p_\star)-H_\Phi(p)\\
    &\qquad
    +\int
    \Bigl[
        \Phi(p_\star)-\Phi(p)
        -\Phi'(p)(p_\star-p)
    \Bigr]\dd x,
\end{align*}
which is the stated entropy--Fisher identity.

Under the assumed NLL response,
$\varphi\in T_p\mathcal M$ implies
\[
    a=A\varphi=\rho\varphi.
\]
Along the reversal path
\[
    p_{\rm rev}(w)=p-wa+o(w),
\]
the derivative of population NLL is
\begin{align*}
    \frac{\dd}{\dd w}
    \E_{p_\star}\!\left[-\log p_{\rm rev}(w)\right]
    \Big|_{w=0}
    &=
    \E_{p_\star}\frac{a}{p}\\
    &=
    \rho\,\E_{p_\star}
    \left(\frac{q}{p}-1\right)\\
    &=
    -\frac{\rho}{Z}
    \bigl(\E_pL-\E_{p_\star}L\bigr).
\end{align*}
Hence reversal decreases model NLL exactly when
$\E_{p_\star}L<\E_pL$.

Finally, if the reweighting is mode seeking, then
Proposition~\ref{prop:mode-seeking-convexity} gives convexity of
$\Phi$, and therefore
$D_\Phi(p_\star\Vert p)\ge0$.  The stated sufficient entropy condition
follows immediately.
\end{proof}
\paragraph{Gaussian trace--determinant relation.}
For centered Gaussian model and target,
\[
    \frac12\left[
        \tr(\Sigma^{-1}\Sigstar)-d
    \right]
    =
    \frac12\log\frac{\det\Sigstar}{\det\Sigma}
    +
    \KL\!\left(
        \mathcal N(0,\Sigstar)\,\middle\|\,
        \mathcal N(0,\Sigma)
    \right).
\]
Thus the NLL trace criterion is the Fisher log-volume criterion plus a
nonnegative KL correction.

\paragraph{Proof of Proposition~\ref{prop:entropy-sampler-dominance}.}
For power tempering,
\[
    q=S_\alpha(p)=\frac{p^\alpha}{Z_\alpha},
    \qquad
    L=p^{\alpha-1},
    \qquad
    Z_\alpha=\E_pL.
\]
Since $q/p=L/Z_\alpha$ and $a=\rho(q-p)$,
\[
    \frac{a}{p}
    =
    \rho\left(\frac{L}{Z_\alpha}-1\right).
\]
Along $p_{\rm rev}(w)=p-wa+o(w)$,
\[
    \frac{\dd}{\dd w}
    \E_{p_\star}\!\left[-\log p_{\rm rev}(w)\right]
    \Big|_{w=0}
    =
    -\frac{\rho}{Z_\alpha}
    \bigl(\E_pL-\E_{p_\star}L\bigr),
\]
which proves (i).

The Fr\'echet derivative of the power-tempering map is
\[
    DS_\alpha|_p[h]
    =
    \alpha q
    \left(
        \frac{h}{p}-\E_q\frac{h}{p}
    \right).
\]
Writing
\[
    S_\alpha(p_{\rm rev}(w))=q-wb+o(w),
    \qquad b=DS_\alpha|_p[a],
\]
gives
\[
    \frac{\dd}{\dd w}
    \E_{p_\star}\!\left[-\log S_\alpha(p_{\rm rev}(w))\right]
    \Big|_{w=0}
    =
    -\frac{\alpha\rho}{Z_\alpha}
    \bigl(\E_qL-\E_{p_\star}L\bigr),
\]
which proves (ii). Finally,
\[
    \E_qL-\E_pL
    =
    \frac{\E_p[L^2]}{Z_\alpha}-Z_\alpha
    =
    \frac{\Var_p(L)}{Z_\alpha}
    \ge0,
\]
which proves (iii).

\subsection{Transparent Fisher sampler-dominance limit}

\begin{corollary}[Transparent Fisher sampler-dominance limit]
\label{cor:sampler-dominance}
Assume $G_J=g_p^F$, $A=\rho\Pi_p^F$, and $B_\kappa=I$ to leading order in
the relevant direction. If
\[
    \|\Pi_p^F\varphi\|_F>\|\Pi_p^F\delta\|_F,
\]
then the favorable sampler component dominates the opposing model component,
so sufficiently small positive reversal weights improve the emitted-sample
objective in this Fisher special case.
\end{corollary}

\begin{proof}
    
 Since $\Pi_p^F$ is self-adjoint and idempotent in the Fisher metric,
\[
    g_p^F(\varphi,\Pi_p^F\varphi)=\|\Pi_p^F\varphi\|_F^2,
\]
and
\[
    g_p^F(\delta,\Pi_p^F\varphi)
      =g_p^F(\Pi_p^F\delta,\Pi_p^F\varphi).
\]
Cauchy--Schwarz then gives
\[
    |g_p^F(\delta,\Pi_p^F\varphi)|
      \le \|\Pi_p^F\delta\|_F\,\|\Pi_p^F\varphi\|_F.
\]
Hence $\|\Pi_p^F\varphi\|_F>\|\Pi_p^F\delta\|_F$ is sufficient for the favorable sampler term to dominate the opposing model term. For an exactly quadratic risk, if $e_0=\delta+\varphi$ and $v=\Pi_p^F\varphi$, improvement along $e_0-w\rho v$ holds on the finite interval
\[
    0<w<\frac{2g_p^F(e_0,v)}{\rho\|v\|_F^2}
\]
whenever $g_p^F(e_0,v)>0$.
\end{proof}

\subsection{Sampler response, reachable ceiling, and residual interpretation}
\label{app:sampler-response}

Writing $S_\kappa(p)=p+\varphi_\kappa(p)$ gives $B_\kappa=I+D\varphi_\kappa(p)$. The full operator need not preserve every direction. In a chosen local sample geometry $G_J$, the relevant requirement is only directional: the reachable sampler-induced displacement should satisfy
\[
    G_J\!\left(\varphi,\,B_\kappa A\varphi\right)>0.
\]
The stronger sampler-dominance condition additionally requires this favorable term to exceed the magnitude of $G_J(\delta,B_\kappa A\varphi)$. No positivity assumption on the full Jacobian is needed.

\paragraph{Reachable ceiling.}
The local emitted correction lies in $\range(B_\kappa\circ A)$. Globally, the relevant family is the composed family $S_\kappa(\mathcal M)$ and the irreducible error is bounded below by $\inf_\theta D(p_\star,S_\kappa(p_\theta))$. The Gaussian toy is degenerate in this respect: with an exact Gaussian score, the DDIM map is affine and maps Gaussian distributions to Gaussian distributions, so the composed family does not leave the Gaussian model class. Real nonlinear models need not share this property.

\paragraph{Inexact residual step.}
The true emitted residual $e_\kappa=\delta+\varphi$ is unavailable without real data. The self-training displacement $A\varphi$ is observable using synthetic data alone. Negative extrapolation therefore uses $A\varphi$ as a surrogate residual and steps against it. Sampler dominance is exactly the condition that this surrogate is aligned with the true emitted residual strongly enough to guarantee improvement.

\section{Experimental implementation details}
\label{app:experimental-details}

\FloatBarrier
\subsection{Scope, reference laws, and independent units}
\label{app:experiment-scope}
The main text is deliberately diffusion-only.  Its experiments correspond to
four distinct components of the local theory: the paired-position study tests
the dependence on the model residual $\delta$; fixed-network comparisons
measure sampler distortion $\varphi$ relative to an operational neutral
reference; baseline subtraction probes the response recorded by training; and
the same-objective plus evaluation-only interventions test the composed
sampler response.  Repeated sampler settings are paired measurements on the
same trained base, not additional independent models.

\begin{table}[htbp]
\centering\small
\caption{Experimental protocols.  These studies use different architectures,
initializations, and reference laws and are not pooled as one seed ensemble.}
\label{tab:experiment-protocols}
\begin{tabularx}{\linewidth}{@{}p{.18\linewidth}p{.18\linewidth}X X@{}}
\toprule
Study & Independent units & Base training and model & Primary role\\
\midrule
Paired position & 30 paired narrow/wide seeds & $N=5000$ per arm; width-64 SiLU residual network with noise skip & Test model-position/trace-side prediction\\
Fixed-model controls & 30 Gaussian + 30 bimodal bases & $N=30000$; four width-32 ReLU blocks; quality-gated checkpoints & Distortion, baseline-subtracted response, same-objective compensation\\
Resolution-only & 30 seeds & $N=5000$; same residual architecture; covariance factor $1.08$ & Hold model/direction fixed and vary evaluation sampler only\\
True-target power pilot & Three seeds per condition; 24 fits & $N=1000,5000$, $d=2,10$; residual architecture & Calibrate score scaling against powers of the true Gaussian\\
Supplementary global-tempering AR & 20 datasets $\times$ three initializations at each $N$ & $N=1000,5000$; two width-64 ReLU layers & Exact-density check of global tempering; appendix only\\
\bottomrule
\end{tabularx}
\end{table}

For a recorded displacement $d$, positive $w$ denotes reversal,
$\theta(w)=\theta_r-wd$.  Local tests differentiate at $w=0$; any fixed
positive weights are separate endpoint checks.  The primary diffusion studies
use actual 50-step Adam displacements.  An exact sampler-free neural density is
not assumed.  Instead, the fixed-model experiments declare the same network
with a 2000-step neutral sampler as the operational reference, and all
conclusions are stated relative to that reference.
\subsection{A paired intervention on model position}
\label{app:paired-position}
This separate study uses 30 paired seeds and a target
$\mathcal N(0,\widetilde\Sigma_\star)$ with eigenvalues $0.5,1.5$ and fixed
QR rotation generated with seed 903. Each seed draws 5000 target observations
once. The narrow and wide training sets multiply these same observations by
$\sqrt{0.8}$ and $\sqrt{1.2}$; the two arms share initialization and random
seeds where applicable. The intervention deliberately separates model
positions; it is not an estimate of naturally occurring error frequencies.

The predictor has three width-64 SiLU hidden layers, 16 sinusoidal time
features, and the parameterization
\[
 \varepsilon_\theta(x,t)=\sqrt{1-\bar\alpha_t}
                   \bigl[x+\bar\alpha_t f_\theta(x,t)\bigr].
\]
The explicit Gaussian-noise skip is an inductive bias absent from the main
plain ReLU model. Base training uses 6000 Adam updates, batch size 256,
cosine learning rate from $10^{-3}$ to $10^{-4}$, and EMA decay 0.995.
The EMA base produces 10000 synthetic observations at source score multiplier
1.1 using 100 DDIM updates. Fine-tuning starts from that base, with fresh
Adam at $10^{-4}$ for 50 updates, batch size 256, and no fine-tuning EMA.

The 1000-index cosine schedule has offset 0.008 and beta cap 0.999. Generation
starts at index 979 from $\mathcal N(0,I)$ and ends at the clean endpoint,
without clipping; initial signal power is about $0.000971$. Unlike the main
fixed-model controls, terminal-prior mismatch remains present. Evaluation
uses neutral score scaling. An independent 16384-observation diagnostic draw
estimates $T=\operatorname{tr}(C_0^{-1}\widetilde\Sigma_\star)/2$. All narrow
intervals lie above one and all wide intervals below one. Mean $T$ is
$1.2901$ (SD $0.0223$) and $0.8632$ (SD $0.0144$), respectively.

The primary objective is Gaussian-moment KL,
\[
 J_G=\tfrac12\!\left[\log\frac{\det C}{\det\widetilde\Sigma_\star}-2
       +\operatorname{tr}(C^{-1}\widetilde\Sigma_\star)+\mu^\top C^{-1}\mu\right].
\]
It includes mean error but is not the full neural-output KL. Secondary
sliced-$W_2^2$ uses 64 projection directions and exact Gaussian target
quantiles. The local derivative is the paired central difference at $h=0.02$,
with $h=0.01,0.04$ as checks. Each model/weight uses eight independent blocks
of 2048 generated points, paired across weights. Across-seed intervals use
30 independent seed pairs; pointwise within-model intervals are not
simultaneous familywise intervals.

\begin{table}[htbp]
\centering\small
\caption{Paired model-position intervention. Brackets are across-seed 95\%
intervals. Counts classify each trained model using pointwise 95\% intervals.}
\label{tab:paired-position}
\begin{tabular}{@{}lcc@{}}
\toprule
Outcome & Narrow arm & Wide arm\\\midrule
Gaussian-moment KL derivative & $-0.04006$ & $0.01635$\\
95\% interval & $[-0.04149,-0.03862]$ & $[0.01572,0.01697]$\\
Denoising-risk derivative & $-0.0033024$ & $0.0026413$\\
95\% interval & $[-0.0034385,-0.0031663]$ & $[0.0025404,0.0027423]$\\
Help / harm / unresolved & 30 / 0 / 0 & 0 / 30 / 0\\\bottomrule
\end{tabular}
\end{table}
The paired wide-minus-narrow KL derivative is $0.05640$, with 95\% interval
$[0.05512,0.05769]$. Direct held-out denoising derivatives use eight independent
8192-example batches and do not involve a reverse-generation sampler.
Doubling evaluation resolution from 100 to 200 preserves all KL and sliced-
Wasserstein point-estimate signs; derivative magnitudes are not invariant
(the maximum relative KL change is 10.5\%). Reversal expands covariance volume
in both arms, but its direction is not purely scalar: reported mean non-scale
fractions are 0.591 and 0.088. Agreement therefore supports the trace-side
prediction in the deliberately separated regimes, not a universal neural
trace criterion or a calibrated boundary near $T=1$.

At the fixed, unselected weight $w=0.25$, mean KL changes are $-0.008728$
with interval $[-0.009063,-0.008393]$ in the narrow arm and $0.004547$ with
interval $[0.004384,0.004711]$ in the wide arm. Sliced-$W_2^2$ changes are
$-0.002077$ and $0.002958$, respectively. Local and finite-weight conclusions
are kept separate. All main seeds 0--29 are retained; engineering pilot
seed 100 is excluded as declared before the main run.

\FloatBarrier
\subsection{Fixed-model diffusion: distortion, response, and compensation}
\label{app:diffusion-protocol}
\label{app:experiment-statistics}
The Gaussian target is $p_\star=\mathcal N(0,\Sigma_\star)$ with $\Sigma_\star=\left(\begin{smallmatrix}2&1\\1&2\end{smallmatrix}\right)$. The non-Gaussian target is
\[
 p_\star=\tfrac12\mathcal N((-2,0),C_0)+\tfrac12\mathcal N((2,0),C_0),
 \qquad C_0=\begin{pmatrix}0.7&0.2\\0.2&1\end{pmatrix}.
\]
The 30 bases per target were trained on 30000 observations each, using a
continuous-time-conditioned noise predictor with linear input/output maps and
four width-32 ReLU blocks. Adam uses batch size 512 and learning rates
$10^{-4}$, $5\times10^{-5}$, and $2\times10^{-5}$ after 0, 10000, and 20000
updates. Candidate checkpoints at 10000, 20000, 30000, and 50000 updates are
accepted by fixed quality gates, not reversal signs: oracle noise-prediction
excess MSE at most 0.005, whitened mean norm at most 0.05, covariance relative
error at most 0.08, and sliced-$W_2^2$ at most 0.005. All 30 bimodal bases pass;
11, 11, and 8 first pass at 10000, 20000, and 30000 updates, respectively.
The controls reuse the retained 60 bases; they are not new base trainings.

Here $K$ counts actual DDIM reverse updates \cite{song2021ddim}, with $K+1$ time-grid points. The
continuous cosine schedule has offset 0.008. Every sampler starts at $t=0.95$
from the exactly noised target law. This known-target initialization isolates
inference effects without claiming robustness to an arbitrary terminal-prior
mismatch. The source settings are $(K,\zeta)\in\{20,500\}\times\{0.9,1.1\}$,
and the reference is the same network at $(2000,1)$. Each source and reference
training run uses 1000 generated observations and 50 fresh-Adam updates at
learning rate $10^{-5}$, with full-batch size 1000. Raw displacements and
paired reference displacements are saved separately.

\paragraph{Fixed-network distortion.}
Let $p_\theta^{\mathrm{ref}}=Q_{2000,1}[s_\theta]$ and
$q_{\theta,\kappa}=Q_{K,\zeta}[s_\theta]$. Keeping $\theta$ fixed, we compare
$q_{\theta,\kappa}$ with $p_\theta^{\mathrm{ref}}$, rather than with $p_\star$.
The first discrepancy is $D_F(P,Q)=\frac12\|\mathbb E_P f-\mathbb E_Q f\|^2$
for a fixed 192-dimensional Fourier feature map. The second is target-whitened
sliced Wasserstein squared, averaged over 64 directions. Gaussian and mixture
target expectations/projection quantiles are computed for the actual target;
the mixture is not replaced by a Gaussian moment approximation.

\begin{table}[htbp]
\centering\small
\caption{Median Fourier distortion relative to the same network's refined
neutral output. Neutral refinement separates discretization from score scaling.}
\label{tab:distortion-controls}
\begin{tabular}{@{}lrrr@{}}
\toprule
Target & Neutral, $K=20$ & Neutral, $K=500$ & $\zeta=1.1$, $K=500$\\\midrule
Gaussian & $4.29525\times10^{-4}$ & $3.76537\times10^{-7}$ & $6.4173\times10^{-3}$\\
Bimodal & $3.89442\times10^{-4}$ & $3.42480\times10^{-7}$ & $4.74562\times10^{-3}$\\\bottomrule
\end{tabular}
\end{table}
The median within-base reduction from neutral $K=20$ to $500$ is 99.9124\%
for Gaussian bases and 99.9127\% for bimodal bases. The corresponding median
gradient-contrast norm reductions are 96.76\% and 96.83\%. All 240 nonneutral
base/source settings have significant positive Fourier distortion after Holm
correction in the distortion-testing family. The exact-target-score version is
an additional oracle diagnostic: at a fixed setting it is the same law for
all base identities, not 30 independently learned oracle models.

\paragraph{Baseline-subtracted response.}
With observations detached during differentiation, let
\[
 \Delta g_\kappa=g(q_{\theta,\kappa})-g(p_\theta^{\mathrm{ref}}),\qquad
 \Delta d_\kappa=d_\kappa-d_{\mathrm{ref}}.
\]
The map from a frozen training law to its initial expected denoising gradient
is linear. Thus $\Delta g_\kappa$ removes the reference-law gradient. The
finite-Adam contrast $\Delta d_\kappa$ is measured, not identified with an
infinitesimal linear response. Tests of nonzero $\mathbb E\Delta g_\kappa$
and positive $-\mathbb E\Delta g_\kappa^\top\Delta d_\kappa$ each survive
separate 240-case Holm corrections in every condition. At $\zeta=1.1$, the
median gradient/endpoint cosines are 0.727 and 0.750 for Gaussian $K=20,500$,
and 0.717 and 0.737 for bimodal $K=20,500$. These are parameter-space cosines.

The median reference/raw endpoint norm ratios at $K=20$, $\zeta=1.1$ are
0.350 and 0.335 for Gaussian and bimodal bases. Norm ratios are not explained-
variance fractions. An oracle contrast $g(q_{\star,\kappa})-g(p_\star)$ has
median cosine 0.996--0.998 with the learned-model contrast across settings.
This subtracts baseline response on both sides; raw-update similarity alone
would not isolate the sampler contribution.

\paragraph{Same-objective test and uncertainty.}
For the original raw $d_\kappa$, differentiate both
$D(p_\star,p_{\theta_r-wd_\kappa}^{\mathrm{ref}})$ and
$D(p_\star,q_{\theta_r-wd_\kappa,\kappa})$ at zero. Each base has eight
independent randomized integration replicas of 2048 points, split into two
independently scrambled Sobol halves. Sampler comparisons share random
coordinates. Cross-products of the independent feature halves give an
unbiased randomized estimator of $D_F$. The empirical Wasserstein estimator
is a finite-sample approximation and is not claimed unbiased.

Pointwise intervals use Student-$t$ quantiles across the eight independent
randomizations, not across individual points within a Sobol net. For the joint
alternative $r_F>0,r_W>0,j_F<0,j_W<0$, take the maximum of the four one-sided
$t$-test $p$-values, then apply Holm correction across all 240 raw base/source
conditions across both targets. The inference is approximate and conditional
on the frozen bases and saved training directions. All bases remain in the
denominator. The main-text counts use this joint endpoint; no single-metric
discovery is substituted for it.

\begin{table}[htbp]
\centering\small
\caption{Ordinary reversal: both deployed objectives improve in all 30 bases
in every row. Counts require reference deterioration and deployed improvement
under both objectives. The maximum adjusted $p$ is over discoveries only.}
\label{tab:full-same-objective}
\begin{tabular}{@{}llrrrr@{}}
\toprule
Target & $K$ & $\zeta$ & Pointwise & Holm & Maximum adjusted $p$\\\midrule
Gaussian & 20 & 0.9 & 2/30 & 2/30 & $3.41\times10^{-5}$\\
Gaussian & 500 & 0.9 & 2/30 & 2/30 & $3.45\times10^{-5}$\\
Gaussian & 20 & 1.1 & 16/30 & 16/30 & $8.42\times10^{-4}$\\
Gaussian & 500 & 1.1 & 16/30 & 16/30 & $6.86\times10^{-4}$\\
Bimodal & 20 & 0.9 & 3/30 & 2/30 & $1.46\times10^{-3}$\\
Bimodal & 500 & 0.9 & 3/30 & 3/30 & $8.07\times10^{-3}$\\
Bimodal & 20 & 1.1 & 16/30 & 16/30 & $2.46\times10^{-3}$\\
Bimodal & 500 & 1.1 & 17/30 & 17/30 & $6.26\times10^{-3}$\\\bottomrule
\end{tabular}
\end{table}
Centered reversal along $\theta-w\Delta d_\kappa$ is a separate diagnostic.
At $\zeta=1.1$ it gives 17/30 Gaussian discoveries at each resolution and
16/30 bimodal discoveries at each resolution, using its own correction
family. It is not counted as ordinary NEON. Denoising-risk results are another
separate objective: the corrected Gaussian campaign gives 11/30 and 13/30
joint denoising/sample discoveries at $\zeta=1.1$, $K=20,500$; the original
bimodal campaign gives 19/30 and 17/30 under its specified testing family.
Those counts neither replace nor add to the same-objective discoveries.

\paragraph{Evaluation-only controls and numerical precision.}
The earlier Gaussian and bimodal source/evaluation grids hold $d_\kappa$
fixed while changing score scaling. Source $\zeta=1.1$ with matched evaluation
improves both sample metrics; opposite-scale $0.9$ evaluation worsens them in
30/30 bases at both displayed resolutions. The Gaussian grid has 35 source
and 35 evaluation settings; the bimodal replication has eight of each. They
are not additional independent bases. A separate 6000-sample, five-replica
bimodal remeasurement confirms the $K=20$ evaluator sign flip in all 30 bases.

The latest randomized-integration reference estimates have median Fourier
standard errors approximately 66 and 59 times smaller than the prior IID
estimates for Gaussian and bimodal bases. The Fourier estimator has the same
unbiased population expectation; for empirical Wasserstein, changing the
point design can also change finite-sample bias. Precision improvement alone
is not proof of equal finite-sample expectations. Across-base mean reference
Fourier effects remain unresolved, so counts of compensation are not evidence
that reversal is harmful on average.

Independent finite-difference checks on six preselected bases at
$h\in\{0.01,0.02,0.04\}$ yield no sign disagreements in 144 objective
comparisons. Reference refinement from 2000 to 4000 updates changes none of
24 tested base/path/metric mean signs; the maximum paired whitened output
RMS is $3.07\times10^{-4}$. This is a selected six-base check, not an all-base
certificate of exact sampling. Source-training replays have small non-bitwise
endpoint differences, all retained in the logs; the primary measurements use
the original saved directions. No density-NLL conclusion follows from these
finite-feature or sample-based discrepancies.

\FloatBarrier
\subsection{Neutral evaluation refinement can remove the apparent gain}
\label{app:resolution-intervention}
This separate 30-seed experiment uses the same target covariance, residual
SiLU architecture, normal-prior initialization, and base/fine-tuning settings
as Appendix~\ref{app:paired-position}, but the 5000-observation training law
has covariance $1.08\widetilde\Sigma_\star$. Source generation remains fixed
at score multiplier 1.1 and 100 DDIM updates, with 10000 synthetic samples
and 50 fine-tuning updates. Neither the base nor the recorded direction is
retrained when evaluation changes.

Evaluation uses score multiplier one throughout, with
$K\in\{10,20,50,100,200,500,980\}$. Each model/weight/step condition has eight
independent 1024-latent blocks, paired across weights and step counts. The
weights are $-0.02,0,0.02,0.25$; $w=0.25$ is fixed, not selected. The main
local outcome is the blockwise central difference of $J_G$ at $h=0.02$.
The same samples give sliced-$W_2^2$ against exact Gaussian projections.

Distortion here is a covariance-level comparison with the same model's fine
reference,
\[
 B_K=\frac{\|\widetilde\Sigma_\star^{-1/2}(C_K-C_{980})
                    \widetilde\Sigma_\star^{-1/2}\|_F}{\sqrt2}.
\]
It is not total error relative to the target. The identity
$C_K-\widetilde\Sigma_\star=(C_{980}-\widetilde\Sigma_\star)+(C_K-C_{980})$
is additive at the matrix level, not after taking norms or KL divergences.
The 980-step reference is the finest unique-index grid in this discrete
schedule, not an exact continuous-time law.

\begin{table}[htbp]
\centering\small
\caption{Evaluation-only resolution intervention. Values are rounded as in
the supplied report. Negative-slope counts and confidence-resolved counts
are distinct; the latter use pointwise 95\% intervals.}
\label{tab:resolution-intervention}
\begin{tabular}{@{}rrrrrr@{}}
\toprule
$K$ & Mean $B_K$ & Mean $T_K$ & \shortstack{Mean local\\KL slope} & \shortstack{Negative\\estimates} & \shortstack{Help / harm /\\unresolved}\\\midrule
10 & 0.2526 & 1.233 & $-0.0301$ & 30/30 & 30 / 0 / 0\\
20 & 0.1265 & 1.067 & $-0.0089$ & 30/30 & 29 / 0 / 1\\
50 & 0.0494 & 0.986 & $0.0013$ & 7/30 & 1 / 5 / 24\\
100 & 0.0235 & 0.962 & $0.0044$ & 0/30 & 0 / 25 / 5\\
200 & 0.0104 & 0.950 & $0.0059$ & 0/30 & 0 / 27 / 3\\
500 & 0.0026 & 0.942 & $0.0068$ & 0/30 & 0 / 29 / 1\\
980 & 0 & 0.940 & $0.0071$ & 0/30 & 0 / 30 / 0\\\bottomrule
\end{tabular}
\end{table}
The exact reported endpoint means are $-0.030147$ with 95\% interval
$[-0.031350,-0.028944]$ at $K=10$ and $0.007053$ with interval
$[0.006302,0.007804]$ at $K=980$. Their within-seed difference averages
$0.037200$, with interval $[0.036523,0.037877]$. The sample-based sliced-
Wasserstein signs also switch from improvement in 30/30 to improvement in
0/30. Covariance distortion decreases and KL slopes increase at every tested
refinement in all 30 models. This is a controlled trajectory across samplers,
not a claim of universal monotonicity in $K$.

Independent 16384-sample fine-reference diagnostics give mean $T=0.9370$
with every within-model interval below one. This is a different draw from
the mean $T_{980}=0.940$ in the evaluation grid. Direct held-out denoising
risk worsens in all 30 models, with mean derivative $0.001101$ and interval
$[0.001008,0.001194]$, unchanged by the evaluator. Thus the coarse evaluator's
contraction can mask the deliberately introduced covariance excess.
The bias factor and grid were chosen using exact-score calculations and an
excluded engineering pilot before the main seeds, not selected from main
outcomes. Near the boundary, pooled versus blockwise moments give 6/30 versus
7/30 negative estimates at $K=50$; the endpoint conclusions are unchanged.
These pointwise intervals are not simultaneous multiple-comparison
guarantees.

\FloatBarrier
\subsection{True-target Gaussian power calibration: a separate pilot}
\label{app:target-power-pilot}
The pilot has four $(N,d)$ settings, $(1000,2),(5000,2),(1000,10),(5000,10)$,
and isotropic and anisotropic target covariances. Three seeds per condition
give 24 models and 72 score-scale evaluations, with seeds paired across
conditions. The isotropic covariance is $I_d$; the anisotropic covariance is
$Q\operatorname{diag}(\operatorname{linspace}(0.5,1.5,d))Q^\top$, with QR
seed $901+d$. Both have trace $d$. This is not the unnormalized covariance
of the primary Gaussian testbed.

The architecture, 6000-update training, EMA, and index-979 normal-prior
sampler match the residual protocol described above. Evaluation uses 100
DDIM updates and $\zeta\in\{0.9,1,1.1\}$. Each condition generates 8192 fit
and 8192 independent test observations. No self-training or reversal is
performed. Fit observations estimate the closest Gaussian power of the
\emph{true} target by
\[
 \widehat\alpha_\star
 =\frac{d}{n^{-1}\sum_i X_i^\top\Sigma_\star^{-1}X_i}.
\]
The uncentered moment includes mean shifts. Independent test covariance
$C$ gives the scale-normalized shape statistic
$\|M/(\operatorname{tr}M/d)-I\|_F/\sqrt d$,
$M=\Sigma_\star^{-1/2}C\Sigma_\star^{-1/2}$. It is not a percentage loss of
sample quality. Because the reference is $p_\star$, this calibration contains
both learned-model error and sampler effects; it does not isolate $q-p_\theta$.

\begin{table}[htbp]
\centering\small
\caption{True-target power pilot. Entries are mean (SD) over three paired
training seeds. Shape entries are the dimensionless residual multiplied by
100. No distributional-equivalence claim follows from these small-seed summaries.}
\label{tab:true-target-power-pilot}
\textit{Fitted target-relative power}\par\smallskip
\begin{tabular}{@{}rrlrrr@{}}
\toprule
$N$&$d$&Covariance&$\zeta=0.9$&$\zeta=1$&$\zeta=1.1$\\\midrule
1000&2&Isotropic&0.512 (0.018)&1.068 (0.026)&2.259 (0.038)\\
1000&2&Anisotropic&0.507 (0.021)&1.077 (0.028)&2.302 (0.042)\\
5000&2&Isotropic&0.489 (0.004)&1.026 (0.009)&2.182 (0.022)\\
5000&2&Anisotropic&0.488 (0.009)&1.040 (0.012)&2.240 (0.034)\\
1000&10&Isotropic&0.492 (0.005)&1.030 (0.011)&2.180 (0.034)\\
1000&10&Anisotropic&0.494 (0.001)&1.036 (0.010)&2.194 (0.037)\\
5000&10&Isotropic&0.485 (0.003)&1.019 (0.008)&2.164 (0.018)\\
5000&10&Anisotropic&0.485 (0.004)&1.026 (0.009)&2.194 (0.021)\\
\bottomrule
\end{tabular}
\par\medskip
\textit{Scale-normalized covariance-shape residual $\times100$}\par\smallskip
\begin{tabular}{@{}rrlrrr@{}}
\toprule
$N$&$d$&Covariance&$\zeta=0.9$&$\zeta=1$&$\zeta=1.1$\\\midrule
1000&2&Isotropic&3.5 (0.4)&3.0 (1.6)&2.9 (2.6)\\
1000&2&Anisotropic&8.2 (2.2)&3.8 (2.0)&7.9 (1.6)\\
5000&2&Isotropic&2.5 (0.5)&2.7 (0.5)&2.8 (0.5)\\
5000&2&Anisotropic&7.2 (1.7)&2.9 (0.9)&5.5 (3.5)\\
1000&10&Isotropic&10.8 (0.7)&11.2 (1.3)&13.1 (1.7)\\
1000&10&Anisotropic&10.2 (0.8)&10.4 (0.2)&14.5 (0.2)\\
5000&10&Isotropic&5.7 (0.2)&6.0 (0.1)&6.3 (0.2)\\
5000&10&Anisotropic&7.7 (0.8)&5.8 (0.5)&6.7 (0.2)\\
\bottomrule
\end{tabular}
\end{table}
Estimated powers need not equal one at neutral scaling. Exact-score
anisotropic controls still have shape residuals about 5.5\% under nonneutral
scaling, while the learned $d=10$, $N=1000$ outputs can have 10--15\% shape
residuals, including substantial neutral error. Standardized sliced-normality
errors range from 0.0147 to 0.0395, versus 0.0166 to 0.0301 in same-size
Gaussian controls; no Gaussianity equivalence test is established. The fitted
power is therefore a useful spread summary, not a proof that the output law
is a power of either the true target or the learned density.

\FloatBarrier

\FloatBarrier
\subsection{Supplementary global-tempering check}
\label{app:ar-global-protocol}
This appendix-only control retains a setting in which both the learned model
and its globally tempered output have explicit densities.  The target is
$p_\star=\mathcal N(0,\left(\begin{smallmatrix}2&1\\1&2\end{smallmatrix}\right))$.
A Gaussian-conditional autoregressive model with two width-64 ReLU layers is
fit on 20 datasets with three initializations at each
$N\in\{1000,5000\}$.  For each frozen model we use the genuine global law
$q_\alpha=p_\theta^\alpha/Z_\alpha$ and deterministic population integration.
At $\alpha=1.10$, reversal improves model NLL in 53/60 fits at $N=1000$ and
55/60 at $N=5000$, while the matched globally tempered output NLL improves in
all 60 $N=5000$ fits.  Hence global sharpening does not by itself determine
model correction.  A same-data refit of only the first marginal reduces the
$N=5000$ helpful count from 55/60 to 45/60, showing that the high favorable
frequency is partly specific to the fitted ensemble rather than a universal
consequence of mode seeking.

The population calculations include the global normalizer and were checked by
8640 direct finite differences across primary and paired controls, with no sign
disagreements.  This study supports the exact power-tempering/NLL discussion,
but it is not needed for the main diffusion mechanism narrative and is not
pooled with the diffusion seed counts.

\FloatBarrier

\end{document}